\documentclass[letterpaper]{article} 
\usepackage{aaai2026}  
\usepackage{times}  
\usepackage{helvet}  
\usepackage{courier}  
\usepackage[hyphens]{url}  
\usepackage{graphicx} 
\usepackage{natbib}  
\usepackage{caption} 
\usepackage{algorithm}
\usepackage{algorithmic}
\usepackage{amsthm,amsmath,amssymb}
\usepackage{mathrsfs}
\usepackage{booktabs}
\usepackage{multirow}
\usepackage{subcaption}
\usepackage{caption}
\usepackage{adjustbox}
\usepackage{colortbl}
\usepackage{xcolor}
\usepackage{caption}
\definecolor{myteal}{RGB}{0, 128, 60} 
\usepackage[hidelinks]{hyperref}
\usepackage{cleveref}
\usepackage{newfloat}
\usepackage{listings}
\DeclareCaptionStyle{ruled}{labelfont=normalfont,labelsep=colon,strut=off} 
\floatstyle{ruled}
\newfloat{listing}{tb}{lst}{}
\floatname{listing}{Listing}
\nocopyright
\title{Exploring Efficient Open-Vocabulary Segmentation in the Remote Sensing}
\author{
    Bingyu Li \textsuperscript{\rm 1, \rm 3}\thanks{Work done during an internship at TeleAI}, 
    Haocheng Dong \textsuperscript{\rm 1, \rm 3}
    Da Zhang\textsuperscript{\rm 2, \rm 3},
    Zhiyuan Zhao\textsuperscript{\rm 3}, 
    Junyu Gao\textsuperscript{\rm 2, \rm 3}, 
    Xuelong Li\textsuperscript{\rm 3}\thanks{Corresponding author}\\
}
\affiliations{
    \textsuperscript{\rm 1}Department of Electronic Engineering and Information Science, University of Science and Technology of China, China \\
    \textsuperscript{\rm 2}School of Artificial Intelligence, OPtics and ElectroNics (iOPEN), Northwestern Polytechnical University, China\\
    \textsuperscript{\rm 3}Institute of Artificial Intelligence (TeleAI), China\\
    \texttt{libingyu0205@mail.ustc.edu.cn, haocheng-dong@mail.ustc.edu.cn, dazhang@mail.nwpu.edu.cn, tuzixini@163.com, gjy3035@gmail.com, xuelong\_li@ieee.org}
}

\usepackage{bibentry}

\begin{document}

\maketitle

Open-Vocabulary Remote Sensing Image Segmentation (OVRSIS), an emerging task that adapts Open-Vocabulary Segmentation (OVS) to the remote sensing (RS) domain, remains underexplored due to the absence of a unified evaluation benchmark and the domain gap between natural and RS images.
To bridge these gaps, we first establish a standardized OVRSIS benchmark (\textbf{OVRSISBench}) based on widely-used RS segmentation datasets, enabling consistent evaluation across methods. Using this benchmark, we comprehensively evaluate several representative OVS/OVRSIS models and reveal their limitations when directly applied to remote sensing scenarios.
Building on these insights, we propose \textbf{RSKT-Seg}, a novel open-vocabulary segmentation framework tailored for remote sensing. RSKT-Seg integrates three key components: (1) a Multi-Directional Cost Map Aggregation (RS-CMA) module that captures rotation-invariant visual cues by computing vision-language cosine similarities across multiple directions; (2) an Efficient Cost Map Fusion (RS-Fusion) transformer, which jointly models spatial and semantic dependencies with a lightweight dimensionality reduction strategy; and (3) a Remote Sensing Knowledge Transfer (RS-Transfer) module that injects pre-trained knowledge and facilitates domain adaptation via enhanced upsampling.
Extensive experiments on the benchmark show that RSKT-Seg consistently outperforms strong OVS baselines by +3.8 mIoU and +5.9 mACC, while achieving 2× faster inference through efficient aggregation. Our code is \href{https://github.com/LiBingyu01/RSKT-Seg}{\textcolor{blue}{here}}.

\section{Introduction}
\label{sec:intro}

Semantic segmentation, a classic task in computer vision, aims to achieve pixel-level category prediction \cite{chen2017deeplab}. Traditional semantic segmentation models are based on manually annotated datasets, which usually cover only a limited number of categories. Similarly, traditional remote sensing image segmentation has long focused on a fixed set of predefined categories \cite{kotaridis2021remote,diakogiannis2020resunet}. However, as application demands grow, especially in scenarios like tracking new urban infrastructure or identifying rare geological features, the shortcomings of this limited-category-based approach have become evident. This has spurred the development of open-vocabulary remote sensing image segmentation (OVRSIS) \cite{ye2025GSNet}.

OVRSIS, builded on the general OVS concept\cite{li2025fgaseg,cho2024cat,xie2024sed,xu2023side}, is tailored to the unique characteristics of remote sensing imagery. By leveraging the power of VLMs and cross-modal learning \cite{radford2021learning,jia2021scaling}, OVRSIS breaks the constraints of traditional training categories, enabling the segmentation of novel classes not present in the original training data. This significantly improves the adaptability and generalization ability of remote sensing image segmentation. 

However, unlike the recent advances in OVS ~\cite{xu2023side, radford2021learning, jia2021scaling}, its extension to the remote sensing (RS) domain (OVRSIS) remains largely underexplored. A critical bottleneck hindering progress in this emerging field is the absence of a standardized evaluation benchmark. Most prior works assess their models on limited datasets or under inconsistent experimental setups, making it difficult to draw fair comparisons or systematically analyze model behavior in remote sensing scenarios.
To address this issue, we construct a unified benchmark for OVRSIS, named \textbf{OVRSISBench}, by reformulating several widely-used remote sensing segmentation datasets under open-vocabulary settings. By leveraging established RS datasets with open-vocabulary configurations, OVRSISBench retains the domain-specific challenges of traditional RS tasks while incorporating the flexibility and generalization demands of open-vocabulary segmentation. This benchmark facilitates fair, consistent, and scalable evaluation of OVS models in realistic RS environments.

\begin{figure}[t]
    \centering
    \includegraphics[width=\linewidth]{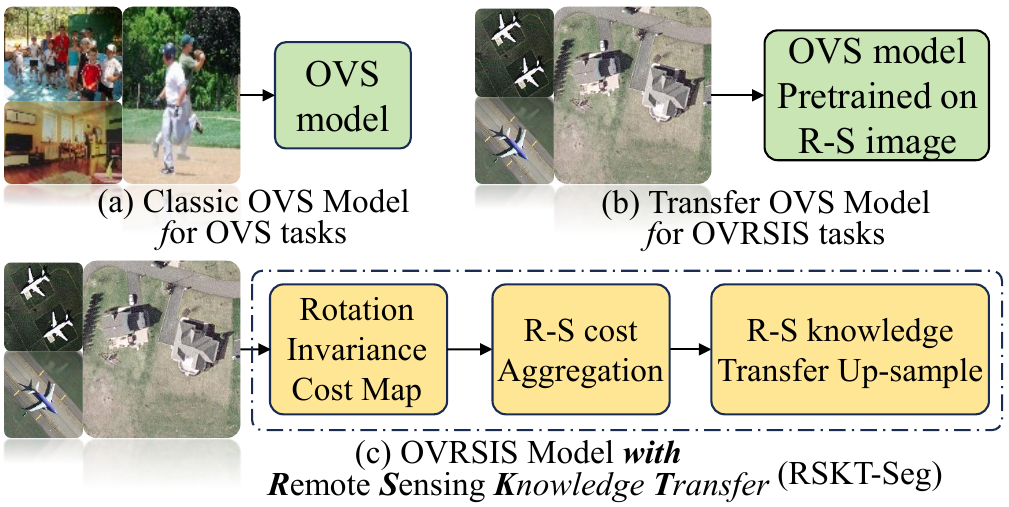}
    \includegraphics[width=\linewidth]{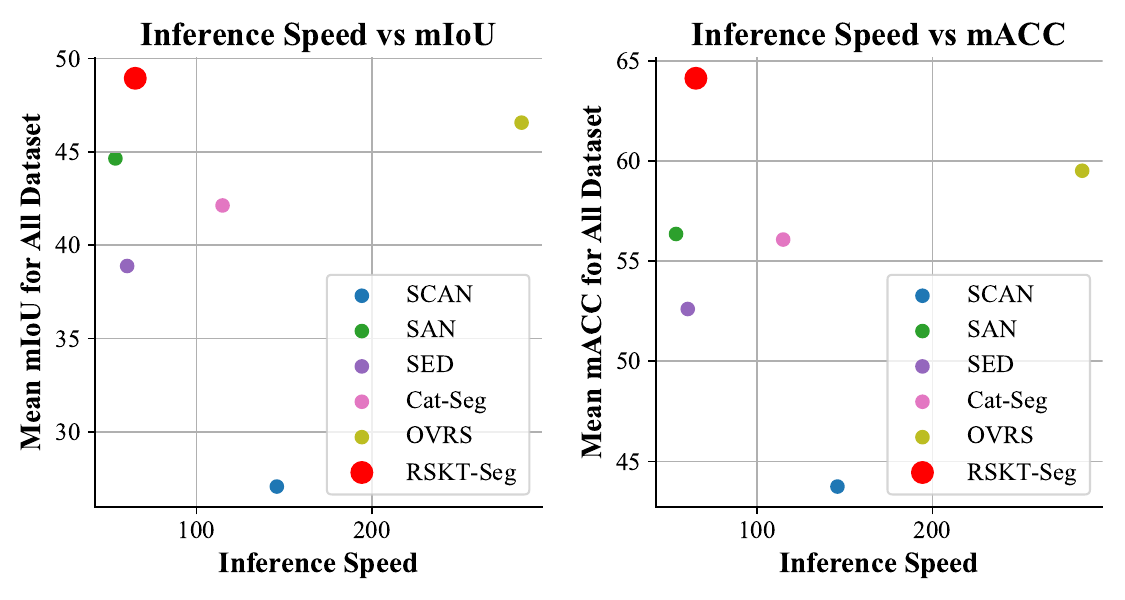}
    \vspace{-20pt}
    \caption{\textbf{(a-c):} Comparison of RSKT-Seg with classic OVS and OVRSIS model. \textbf{(d):} Comparison of RSKT-Seg with different models in terms of inference speed against mean Intersection over Union (mIoU) on the left and against mean Accuracy (mACC) on the right.}
    \label{fig:fig_01}
    \vspace{-10pt}
\end{figure}

Using this \textbf{OVRSISBench} benchmark, we comprehensively evaluate several representative OVS models and observe a significant drop in performance when these models are directly transferred from natural images to the RS domain (see Fig.~\cref{fig:fig_01}(a-b)). 
Furthermore, we survey recent works on OVRSIS model and evaluate them under the proposed \textbf{OVRSISBench}. Our experiments reveal that while some approaches demonstrate incremental improvements compared with OVS model, they often fail to achieve effective knowledge transfer to the RS domain. For example, OVRS~\cite{cao2024open}, which introduces only minor modifications to OVS model \cite{cho2024cat}, shows performance gains but lacks mechanisms to incorporate RS-specific priors, leading to limited adaptability. Similarly, models such as GSNet~\cite{ye2025GSNet}, although partially incorporating RS knowledge, overlook key characteristics of RS imagery such as large-scale context, object rotation, and spectral diversity.

Overall, these findings highlight two key limitations of existing OVS and OVRSIS approaches: 
(1) classic OVS models exhibit limited transferability when applied directly to open-vocabulary tasks in the remote sensing domain; 
(2) existing OVRSIS methods fail to adequately model remote sensing-specific characteristics such as rotation invariance and large-scale spatial context.

To fill these gaps, we propose \textbf{RSKT-Seg}, an efficient framework for open-vocabulary segmentation in remote sensing that achieves both high segmentation accuracy and fast inference speed (see Fig.~\cref{fig:fig_01}(d)). First, we introduce a \textit{Multi-Direction Remote Sensing Cost Map Aggregation} (RS-CMA) module to capture the rotation-invariant characteristics of RS images by computing vision-text similarities from multiple directions. Second, we design an efficient \textit{Cost Map Fusion} (RS-Fusion) strategy that simultaneously considers spatial and class-wise interactions, while incorporating a dimensionality reduction mechanism to accelerate inference without sacrificing performance. Finally, we propose a \textit{Remote Sensing Knowledge Transfer} (RS-Transfer) upsampling module that leverages pre-trained model knowledge to facilitate effective domain adaptation to RS imagery.

\noindent\textbf{Our main contributions are summarized as follows:}
\begin{itemize}
    \item \textbf{Benchmark.} We construct \textbf{OVRSISBench}, an unified benchmark for open-vocabulary remote sensing image segmentation.
    \item \textbf{Evaluation.} Based on OVRSISBench, we comprehensively evaluate representative OVS models and recent OVRSIS methods. Our analysis reveals their potential limitations and provides a standardized reference for future research.
    \item \textbf{Framework.} We propose \textbf{RSKT-Seg}, an efficient and effective framework for open-vocabulary segmentation in remote sensing, which achieves both high segmentation accuracy and fast inference speed (shown in \cref{fig:fig_01}(d)).
\end{itemize}
    
\section{Related Works}
\label{sec:related_works}

\subsection{Semantic Segmentation}
Semantic segmentation, a crucial task in computer vision, focuses on pixel-level classification. Over the years, it has seen significant progress with various methods and models.
Fully convolutional networks (FCNs) \cite{long2015fully} were an early milestone. As end-to-end models, they enabled direct pixel-wise predictions. Later, SegNet \cite{badrinarayanan2017segnet} and U-Net \cite{ronneberger2015u} evolved from FCNs. These encoder-decoder architectures effectively captured low-level and high-level features, gaining popularity. ResNet is widely used as a feature encoder in semantic segmentation, forming the basis of many models \cite{lin2018deeptongue}. 
Recently, Vision Transformers (ViT) \cite{dosovitskiy2020image} have brought new ideas. By using self-attention, ViT models capture long-range dependencies. Adaptations like SETR \cite{zheng2021rethinking} improve efficiency. New paradigms based on ViT have emerged for semantic segmentation \cite{li2024u3m, hu2024contrastive}, which requires precise pixel-level feature handling \cite{xie2024multi, zhu2024saswot, li2024stitchfusion}.
Notable works such as MaskFormer \cite{cheng2021per} and Mask2Former \cite{cheng2022masked} unify pixel-level and mask classification, enhancing performance. SegFormer addresses resolution issues by removing positional encodings and presenting multi-resolution features \cite{xie2021segformer}. We summarize works about OVS in appendix A.

\begin{figure*}[t]
    \centering
    \includegraphics[width=0.88\linewidth]{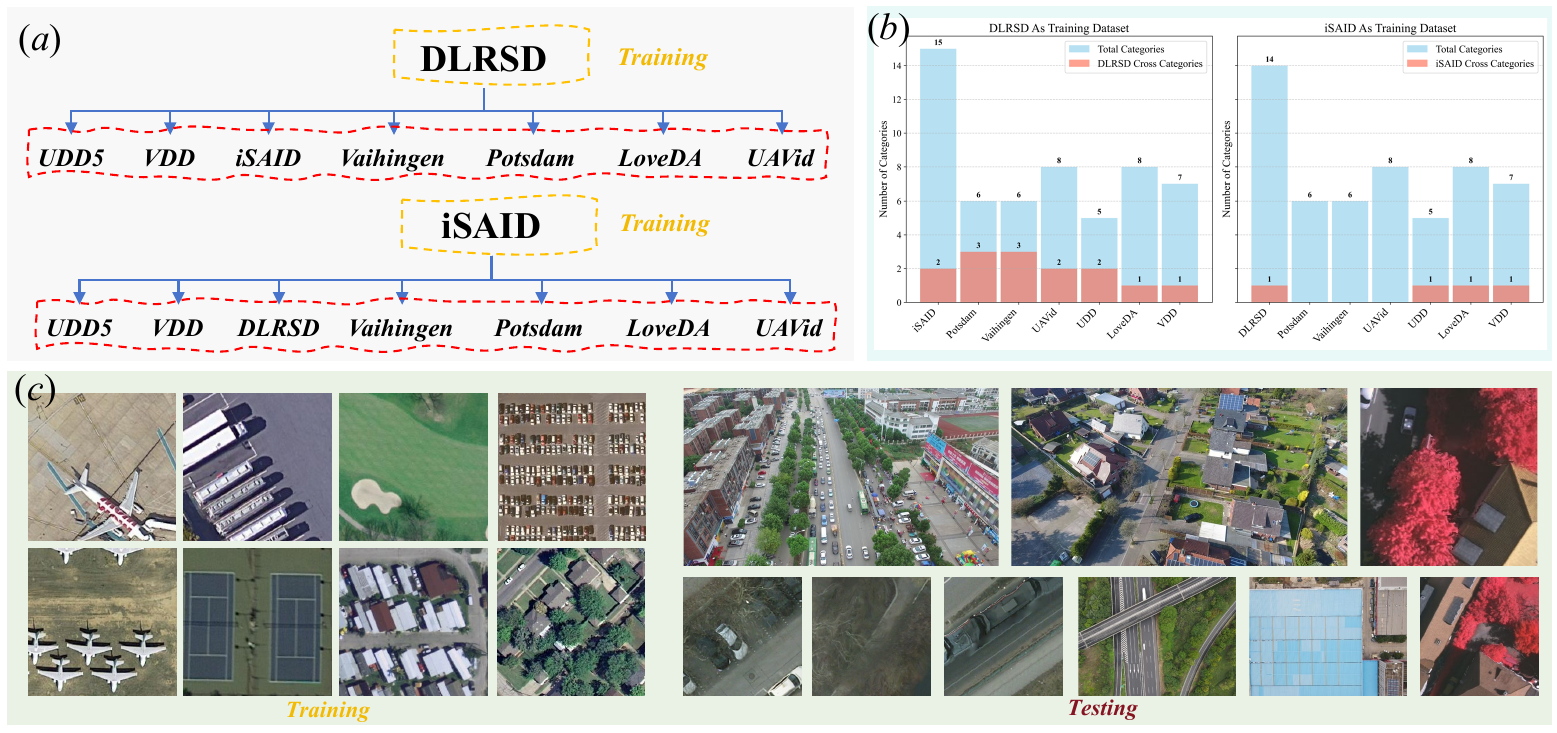}
    \vspace{-8pt}
    \caption{\textbf{Schematic diagram of OVRSISBench} (a) Dataset division based on the open-vocabulary protocol (b) Vocabulary (class) overlap number between training and test datasets under two division scenarios (c) Examples display of training and test sets. The more information is in the appendix.}
    \vspace{-8pt}
    \label{fig:fig_2_dataset}
\end{figure*}

\subsection{Open-Vocabulary Remote Sensing Image Segmentation}
OVRSIS is an emerging field in a rapid development stage, and preliminary exploratory research work has been carried out. In the process of migrating from the OVS field to the remote sensing field, OVRS \cite{cao2024open} takes into account the basic characteristics of the remote sensing field, makes some fundamental improvements on catseg, and has achieved certain results. GSNet \cite{ye2025GSNet}introduces a specialist-generalist model, which can better integrate remote sensing domain knowledge into the model. SegEarth-OV \cite{li2024segearth} proposes a train-free model, which shows better performance than the OVS model on 17 datasets. Overall, as an emerging field, in OVRSIS, few existing models can consider remote sensing features and introduce remote sensing domain knowledge for migration. Most of them only modify and improve the OVS model. In view of this, we propose the RSKT-Seg model. By considering the rotational invariance of remote sensing images and introducing remote sensing domain knowledge, this model has obtained satisfactory performance on multiple datasets. 

\section{Benchmark: OVRSISBench}
\label{sec:OVRSISBench}

To promote standardized evaluation in the emerging field of open-vocabulary remote sensing image segmentation (OVRSIS), we construct \textbf{OVRSISBench}, a unified and scalable benchmark that reformulates several widely-used remote sensing datasets under an open-vocabulary setting.

\subsection{Benchmark Construction}

\subsubsection{Dataset Construction}
\textbf{OVRSISBench} follows an open-vocabulary paradigm like \cite{cao2024open}, where the model must generalize to previously unseen classes based on text descriptions (\cref{fig:fig_2_dataset}(a)). To construct this benchmark, we adapt 8 representative remote sensing datasets: DLRSD~\cite{chaudhuri2017multilabel}, iSAID~\cite{yao2021scale}, Potsdam, Vaihingen, UAVid\cite{LYU2020108}, UDD5\cite{chen2018large}, LoveDA\cite{NEURIPS_DATASETS_AND_BENCHMARKS2021_4e732ced} and VDD\cite{cai2025vdd}. These datasets cover a diverse range of scenes, including urban layouts, agricultural regions, and high-resolution aerial imagery, the examples of the datasets are illustrated in the \cref{fig:fig_2_dataset}(c). We introduce the datasets in the appendix C. 

\subsubsection{Dataset Usage and Splits}
As shown in \cref{fig:fig_2_dataset}(a), we utilize DLRSD and iSAID as training sets as \cite{cao2024open} due to their large scale and diversity, enabling robust learning of remote sensing visual patterns. For evaluation, we test across all 8 datasets to ensure the model's generalization under varying scene distributions and resolutions. The training and evaluation splits are consistent across all methods to ensure fair comparison.

\subsection{Open-Vocabulary Protocol Analysis}
To rigorously assess the open-vocabulary generalization capability of the models, particularly its performance on unseen classes. We adopt a cross-dataset transfer protocol in our \textbf{OVRSISBench}. In this setting, the models are trained on the DLRSD and iSAID datasets and evaluated on the separate target datasets with partially disjoint category sets. This protocol aligns with the standard open-vocabulary segmentation (OVS) setting\cite{xu2022simple, cho2024cat}. As shown in \cref{fig:fig_2_dataset}(b), we further perform a statistical analysis of the cross-category distribution to substantiate the open-vocabulary nature of our setup.  

\begin{figure*}[t]
    \centering
    \includegraphics[width=0.92\linewidth]{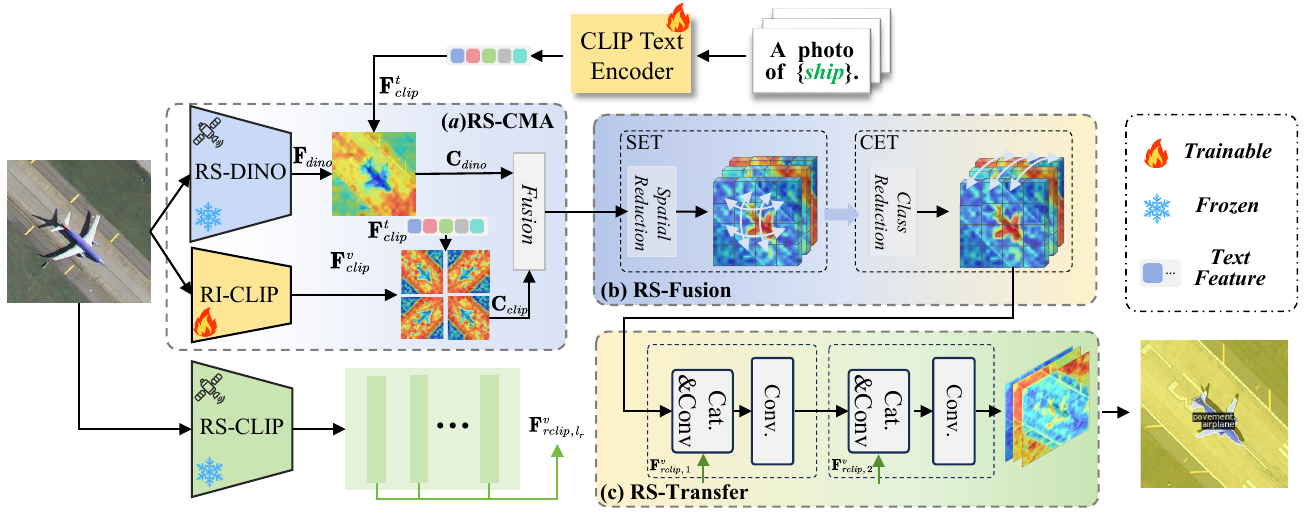}
    \vspace{-8pt}
    \caption{\textbf{The overall framework of RSKT-Seg} includes: (a) the overall procedure of RS-CMA module; (b) the workflow of the RS-Fusion Module; (c) the framework of the RS-Transfer Upsample. The more detailed framework is in appendix J}
    \vspace{-8pt}
    \label{fig:fig_02}
\end{figure*}

\subsection{Evaluation on OVRSISBench}

\subsubsection{Model Selection and Experiments}
We first evaluated the performance of several classic OVS model on \textbf{OVRSISBench}, and subsequently selected recently published OVRSIS methods for evaluation. 
For the evaluation metrics, we adopted mean Intersection over Union (mIoU) and mean Accuracy (mACC). We introduce the Evaluation metric in the appendix D. 
We utilized the pre-trained ViT-B/16 \cite{dosovitskiy2020image} and ViT-L/14@336 \cite{radford2021learning} as the vision-language models.
\subsubsection{Analysis of Existing Model Limitations}
Table \ref{tab:01_main_table} shows classic OVS methods and existing OVRSIS methods perform poorly overall, with clear room for improvement.
Classic OVS methods, designed for natural scenes, lag behind dedicated OVRSIS methods under the same backbone.
In contrast, although existing OVRSIS methods perform better than classic OVS methods, they still have limitations in fully exploiting remote sensing domain knowledge.

\section{Method: RSKT-Seg}
\label{sec:method}
Although previous OVRSIS models have achieved certain advancements, as we analyzed in the introduction section, we attribute the low accuracy observed in the tests on OVRSISBench to the lack of effective transfer of remote sensing domain knowledge. Therefore, we have designed the \textbf{RSKT-Seg} with remote sensing domain knowledge transfer capabilities.

\subsection{Overall Architecture of RSKT-Seg}
The overall framework of our proposed RSKT-Seg is shown in Figure \cref{fig:fig_02}(a). We design the OVRSIS architecture with remote sensing knowledge transfer.
The preparatory knowledge for training pre-trained VLMs such as CLIP has been extensively introduced in previous methods \cite{hu2024contrastive}. The definition of OVRSIS is the same as that of OVS. We will not elaborate on it here, and researchers can refer to the corresponding sections of relevant literature \cite{li2025fgaseg,cho2024cat,xu2023side}. 

\begin{figure}[t]
    \centering
    \includegraphics[width=0.95\linewidth]{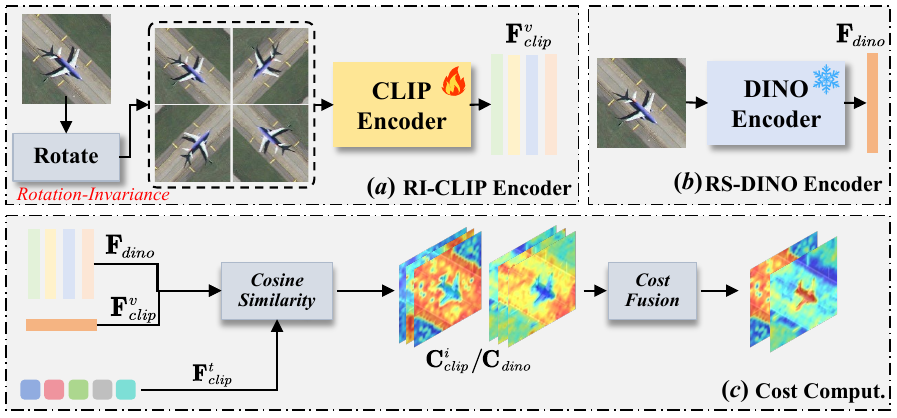}
    \vspace{-8pt}
    \caption{\textbf{(a)} Multi-rotation feature encoding using CLIP \textbf{(b)} feature encoding using RS-DINO and \textbf{(c)} cost map construction using CLIP and DINO. These modules do not introduce learnable parameters.}
    \label{fig:fig_03}
    \vspace{-15pt}
\end{figure}


\subsection{Remote Sensing Cost Map Aggregation (RS-CMA)}
As shown in \cref{fig:fig_2_dataset}(c), unlike natural images, remote sensing (RS) imagery often captures scenes from top-down perspectives, where object orientation is arbitrary due to aerial rotation or satellite orbit variations. 
Consequently, the same semantic class (e.g., ``bridge'', ``airplane'') may appear under drastically different orientations across samples. This intrinsic variability poses a significant challenge for open-vocabulary recognition. To address this, we explicitly introduce \textbf{rotation-invariance} through our \textit{Multi-Direction Remote Sensing Cost Map Aggregation (RS-CMA)} module.

Specifically, we augment the input image \(\mathbf{I}\) with the height \(H\) and width \(W\) in four directions, obtaining \(\mathbf{I}_i\) (\(i = 0,1,2,3\)) with the same shape (shown in \cref{fig:fig_03}(a)). Among them, $\mathbf{I}_0$ is the original orientation, and 1, 2, 3 respectively represent the enhanced images in the orientations rotated by $90 \times i$. 

As \cref{fig:fig_03}(a-b), we use the CLIP image encoder \(\mathcal{E}_{clip}^v\) to encode \(\mathbf{I}_i\), yielding \(\mathbf{F}_{clip}^{v,i} = \mathcal{E}_{clip}^v(\mathbf{I}_i)\).
To embed remote-sensing domain knowledge, we employ a pre-trained DINO encoder \(\mathcal{E}_{dino}\) on a large-scale dataset \cite{ye2025GSNet}. Encoding \(\mathbf{I}_0\) with \(\mathcal{E}_{dino}\), we get \(\mathbf{F}_{dino}=\mathcal{E}_{dino}(\mathbf{I}_0)\).

Given a set of classes \(\mathcal{C} \in \mathbb{R}^{N_t}\), we obtain text embeddings \(\mathbf{F}^{t}_{clip}=\mathcal{E}_{clip}^t(\mathcal{C}) \in \mathbb{R}^{N_t \times C_f}\). Then, for each \(\mathbf{I}_i\), we compute a cost volume \(\mathbf{C}_{clip}^i \in\mathbb{R}^{(H_f\times W_f)\times N_t}\) using cosine similarity:
\begin{equation}
\mathbf{C}_{clip}^i(j, n)=\frac{\mathbf{F}_{clip}^{v,i}(j)\cdot\mathbf{F}^{t}_{clip}(n)}{\|\mathbf{F}_{clip}^{v,i}(j)\|\|\mathbf{F}^{t}_{clip}(n)\|}
\end{equation}
where \(j\) represents 2D spatial positions of the image embedding and \(n\) is an index for a class.

Similarly, for the DINO-based features, we compute a cost volume \(\mathbf{C}_{dino}\in\mathbb{R}^{(H_f\times W_f)\times N_t}\) for \(\mathbf{I}\) as:
\begin{equation}
\mathbf{C}_{dino}(j, n)=\frac{\mathbf{F}_{dino}(j)\cdot\mathbf{F}^{t}_{clip}(n)}{\|\mathbf{F}_{dino}(j)\|\|\mathbf{F}^{t}_{clip}(n)\|}
\end{equation}

Next, we apply a cost map fusion function across the four rotated variants and the DINO-based cost map to generate a rotation-invariant and domain-aware fused cost map (as shown in \cref{fig:fig_03}(c)):
\begin{equation}
\mathbf{C}_{fused} = Fusion(\mathbf{C}_{clip}^i, \mathbf{C}_{dino}), \quad i=0,1,2,3
\end{equation}
The fused cost map is further organized under multiple prompt templates, forming a tensor of shape $H_f \times W_f \times N_t \times P$, where $P$ denotes the number of templates. $\mathbf{C}_{fused}$ is projected into $\mathbf{C}_{s}$ with the shap $H_f \times W_f \times N_t \times C_f$ by a linear layer.

\subsection{Efficient Cost Map Fusion (RS-Fusion)}
The cost map $\mathbf{C}_{s}$ is used to characterize the correlation between visual features and text features. Building upon this, we gradually predict the segmentation map by enhancing the cost map's \textbf{spatial discriminative ability} (\(H_f W_f\) dimension) and \textbf{class discriminative ability} (\(N_t\) dimension). To achieve this, we have designed two dedicated modules: the Spatial Enhancement Transformer (SET) and the Class Enhancement Transformer (CET), as illustrated in \cref{fig:fig_02}(b) and (c). 
This idea aligns with that presented in \cite{cho2024cat}. To further strengthen the integration of remote sensing domain knowledge, we embed knowledge from the pre-trained DINO model together with the cost map. Additionally, considering the requirement for faster inference speed, we have developed two feature dimension reduction methods and a lightweight transformer architecture.

\subsubsection{Spatial Enhancement Transformer(SET)}
To enhance the semantic representation with richer spatial information, we concatenate the cost map with intermediate-level features from both CLIP and DINO. These features are first projected to a unified dimension and then passed through a convolutional layer \(\mathcal{R}_s\) that reduces the spatial resolution, yielding a compressed tensor for fast inference (as shown in \cref{fig:fig_02}(b)):
\begin{equation}
\mathbf{C}_{sr} = \mathcal{R}_s([\mathbf{C}_s; \mathbf{F}_{clip}^{v}; \mathbf{F}_{dino}])
\end{equation}
Here, $\mathcal{R}_s$ denotes the spatial reduction module and \([\cdot;\cdot;\cdot]\) is concatenate operation. Then the reduced feature \(\mathbf{C}_{sr}\) is used as key and value in a Transformer block designed to aggregate spatial context via cross-attention manner:
\begin{equation}
\mathbf{C}_{so} = \text{Transformer}_s(\mathbf{C}_s, \mathbf{C}_{sr}, \mathbf{C}_{sr})
\end{equation}

\subsubsection{Class Enhancement Transformer(CET)}
As shown in \cref{fig:fig_02}(b), we further refine the fused features \(\mathbf{C}_{so}\) in the class dimension by integrating the text features \(\mathbf{F}^t_{clip}\) from CLIP. After applying average pooling $Avg(\cdot)$ to downsample the spatial resolution:
\begin{equation}
\mathbf{C}_{cr} = \text{Avg}([\mathbf{C}_{so}; \mathbf{F}^t_{clip}]),
\end{equation}
we reshape the cost map to match the shape of repeated text features. The two are concatenated and passed through a second Transformer block to process the class dimension:
\begin{equation}
\mathbf{C}_{co} = \text{Transformer}_c(\mathbf{C}_{cr}, \mathbf{C}_{cr}, \mathbf{C}_{cr})
\end{equation}
This allows the model to capture interactions across different categories. 

Through iterative processing across \(N\) layers, the aforementioned SET and CET enhance the features of the cost map in both spatial and class dimensions. This strengthens the discriminability of the cost map for objects belonging to different categories as well as for objects within the same class, ultimately forming the fused cost map $\mathbf{C}_{agg}$.
The analysis for Dimension Reduction is in the appendix B.

\subsection{RS-Transfer Upsample}

The fused cost map $\mathbf{C}_{agg}$ provides spatial and class-discriminative signals but lacks detailed textures due to the low spatial resolution. To address this, as shown in \cref{fig:fig_02}(c), we integrate intermediate-layer features from RemoteCLIP vision encoder  $\mathbf{F}_{rclip}^v$, CLIP vision encoder $\mathbf{F}_{clip}^v$, and DINO encoder $\mathbf{F}_{dino,l_d}$ during the upsampling process. We concatenate these features with the upsampled cost map $\mathbf{C}_{agg}$ and pass them through a projection module:
\begin{equation}
\mathbf{C}_{agg} = \text{Proj}([\mathbf{C}_{agg}; \mathbf{F}_{rclip,l_r}^v; \mathbf{F}_{clip,l}^v; \mathbf{F}_{dino,l_d}])
\end{equation}
Here, $\text{Proj}$ is a feature projection module. After $N_d$ layers of refinement, the final prediction $\text{O}$ is generated in the original image resolution, with shape $N_t \times H \times W$.

\begin{table*}[htbp]
    \centering
    \setlength{\tabcolsep}{2.8pt} 
    \caption{\textbf{Main Results of the RSKT-Seg on 8 OVRSIS Datasets.} Evaluation metrics: mIoU, mACC. Best results are bold, second-best underlined. m-mIoU and m-mACC denote average values across all datasets.}
    \label{tab:01_main_table}
    \vspace{-10pt}
    \begin{adjustbox}{max width=0.92\textwidth}
    \begin{tabular}{lllcccccccccccccccccc}
        \toprule
        \multicolumn{21}{c}{DLRSD as Training Dataset} \\
        \midrule
        \midrule
        \multirow{2}*{\textbf{Method}} & \multirow{2}*{\textbf{Backbone}} & \multirow{2}*{\textbf{Type}} & \multicolumn{2}{c}{DLRSD} & \multicolumn{2}{c}{iSAID} & \multicolumn{2}{c}{LoveDA} & \multicolumn{2}{c}{Potsdam} & \multicolumn{2}{c}{UAVid} & \multicolumn{2}{c}{UDD5} & \multicolumn{2}{c}{Vaihingen} & \multicolumn{2}{c}{VDD} & \multicolumn{2}{c}{Mean of All Datasets}\\ 
        \cmidrule(lr){4-5} \cmidrule(lr){6-7} \cmidrule(lr){8-9} \cmidrule(lr){10-11} \cmidrule(lr){12-13} \cmidrule(lr){14-15} \cmidrule(lr){16-17} \cmidrule(lr){18-19} \cmidrule(lr){20-21}
        & & & mIoU & mACC & mIoU & mACC & mIoU & mACC & mIoU & mACC & mIoU & mACC & mIoU & mACC & mIoU & mACC & mIoU & mACC & m-mIoU & m-mACC\\ 
        \midrule
        SCAN$_{\textcolor{gray}{\text{\tiny \textit{CVPR2024}}}}$ & ViT-B & OVS & 48.52 & 68.68 & 34.18 & 49.05 & 18.23 & 40.12 & 20.22 & 34.70 & 18.56 & 30.31 & 32.12 & 40.45 & 5.38 & 22.54 & 26.25 & 43.67 & 25.43 & 41.19 \\
        SAN$_{\textcolor{gray}{\text{\tiny \textit{CVPR2023}}}}$ & ViT-B & OVS & 85.73 & 91.03 & 30.63 & 44.03 & 23.15 & 48.26 & \underline{30.30} & \underline{44.98} & 22.34 & 37.56 & 36.87 & 47.21 & 31.92 & 45.36 & 34.76 & 52.42 & 36.96 & 51.36 \\
        SED$_{\textcolor{gray}{\text{\tiny\textit{CVPR2024}}}}$ & ConvNeXt-B & OVS & 85.13 & 91.36 & 21.54 & 36.28 & 21.32 & 45.17 & 19.47 & 33.40 & 20.12 & 33.34 & 34.65 & 44.10 & 29.40 & 49.38 & 31.43 & 50.25 & 32.88 & 47.91 \\
        Cat-Seg$_{\textcolor{gray}{\text{\tiny\textit{CVPR2024}}}}$ & ViT-B & OVS & 85.84 & 91.44 & 23.56 & 38.48 & 25.45 & 50.32 & 26.79 & 44.72 & 24.56 & 39.20 & 38.23 & 49.56 & 32.32 & 49.65 & 36.18 & 54.30 & 36.62 & 52.21 \\
        OVRS$_{\textcolor{gray}{\text{\tiny\textit{TGRS2025}}}}$ & ViT-B & OVRSIS & \underline{85.98} & \underline{91.52} & 39.09 & 54.43 & 28.67 & 52.10 & 27.47 & 42.07 & 25.23 & 40.18 & 39.10 & 50.65 & 33.71 & 49.01 & 37.34 & 55.15 & 39.57 & 54.51 \\
        GSNet$_{\textcolor{gray}{\text{\tiny\textit{AAAI2025}}}}$ & ViT-B & OVRSIS & 84.12 & 90.53 & \underline{42.00} & \underline{59.19} & \underline{29.32} & \underline{53.02} & 26.46 & 43.20 & \underline{25.42} & \underline{40.70} & \underline{40.05} & \underline{51.72} & \underline{35.15} & \underline{52.62} & \underline{38.10} & \underline{56.01} & \underline{40.08} & \underline{55.87} \\
        
        RSKT-Seg & ViT-B & OVRSIS & \textbf{90.60} & \textbf{94.89} & \textbf{44.04} & \textbf{61.23} & \textbf{32.49} & \textbf{55.67} & \textbf{34.53} & \textbf{50.71} & \textbf{28.14} & \textbf{42.86} & \textbf{42.99} & \textbf{57.81} & \textbf{37.16} & \textbf{54.81} & \textbf{41.22} & \textbf{58.09} & \textbf{43.90} & \textbf{59.51} \\
        \midrule
        SCAN$_{\textcolor{gray}{\text{\tiny \textit{CVPR2024}}}}$ & ViT-L & OVS & 52.42 & 72.43 & 44.28 & 67.25 & 23.17 & 35.36 & 27.45 & 39.22 & 20.28 & 34.43 & 34.14 & 43.25 & 15.23 & 29.45 & 29.24 & 45.57 & 30.78 & 45.87 \\
        SAN$_{\textcolor{gray}{\text{\tiny \textit{CVPR2023}}}}$ & ViT-L & OVS & 86.45 & 91.25 & 49.56 & 67.25 & 25.33 & 37.54 & 37.25 & 46.28 & 23.53 & 38.14 & 37.23 & 48.45 & 39.22 & 48.33 & 35.83 & 53.25 & 41.80 & 53.81 \\
        SED$_{\textcolor{gray}{\text{\tiny \textit{CVPR2024}}}}$ & ConvNeXt-L & OVS & 87.68 & 91.24 & 51.23 & 68.24 & 24.55 & 36.83 & 29.35 & 37.95 & 21.33 & 35.64 & 35.73 & 45.15 & 39.02 & 58.62 & 32.53 & 51.34 & 40.18 & 53.13 \\
        Cat-Seg$_{\textcolor{gray}{\text{\tiny \textit{CVPR2024}}}}$ & ViT-L & OVS & 88.68 & 93.34 & 53.34 & 70.86 & 28.64 & 38.73 & 35.75 & 49.03 & 25.73 & \underline{40.54} & 40.24 & 51.65 & 42.30 & 60.65 & \underline{39.14} & 55.85 & 44.23 & 57.58 \\
        OVRS$_{\textcolor{gray}{\text{\tiny\textit{TGRS2025}}}}$ & ViT-L & OVRSIS & \underline{88.85} & \underline{93.64} & 52.65 & 69.59 & 31.53 & \underline{59.82} & 36.44 & 50.17 & 24.13 & 34.83 & 40.82 & 54.24 & \underline{43.50} & \textbf{63.31} & 37.23 & 56.34 & 44.39 & 60.24 \\
        GSNet$_{\textcolor{gray}{\text{\tiny\textit{AAAI2025}}}}$ & ViT-L & OVRSIS & 86.02 & 91.48 & \underline{53.73} & \underline{71.57} & \underline{32.52} & \textbf{60.23} & \underline{37.85} & \underline{52.35} & \underline{24.22} & 35.03 & \underline{40.92} & \underline{57.04} & \textbf{44.13} & 62.38 & 37.34 & 57.04 & \underline{44.56} & \underline{60.89} \\
        RSKT-Seg & ViT-L & OVRSIS & \textbf{93.49} & \textbf{96.49} & \textbf{54.32} & \textbf{71.72} & \textbf{33.23} & 57.41 & \textbf{38.44} & \textbf{56.69} & \underline{25.72} & \textbf{40.89} & \textbf{42.10} & \underline{56.39} & 42.69 & \underline{63.29} & \textbf{39.69} & \textbf{58.60} & \textbf{46.21} & \textbf{62.69} \\
        \bottomrule


        \toprule
        \multicolumn{21}{c}{iSAID as Training Dataset} \\
        \midrule
        \midrule
        \multirow{2}*{\textbf{Method}} & \multirow{2}*{\textbf{Backbone}} & \multirow{2}*{\textbf{Type}} & \multicolumn{2}{c}{DLRSD} & \multicolumn{2}{c}{iSAID} & \multicolumn{2}{c}{LoveDA} & \multicolumn{2}{c}{Potsdam} & \multicolumn{2}{c}{UAVid} & \multicolumn{2}{c}{UDD5} & \multicolumn{2}{c}{Vaihingen} & \multicolumn{2}{c}{VDD} & \multicolumn{2}{c}{Mean of All Datasets}\\ 
        \cmidrule(lr){4-5} \cmidrule(lr){6-7} \cmidrule(lr){8-9} \cmidrule(lr){10-11} \cmidrule(lr){12-13} \cmidrule(lr){14-15} \cmidrule(lr){16-17} \cmidrule(lr){18-19} \cmidrule(lr){20-21}
        & & & mIoU & mACC & mIoU & mACC & mIoU & mACC & mIoU & mACC & mIoU & mACC & mIoU & mACC & mIoU & mACC & mIoU & mACC & m-mIoU & m-mACC\\ 
        \midrule
        SCAN$_{\textcolor{gray}{\text{\tiny \textit{CVPR2024}}}}$ & ViT-B & OVS & 16.09 & 38.25 & 62.34 & 76.48 & 12.56 & 32.10 & \underline{18.25} & 33.17 & 9.87 & 15.30 & 8.45 & 28.60 & 8.72 & 27.20 & 15.32 & 35.10 & 18.95 & 35.78 \\
        SAN$_{\textcolor{gray}{\text{\tiny \textit{CVPR2023}}}}$ & ViT-B & OVS & 18.82 & 42.36 & 85.43 & 90.36 & 18.45 & 40.25 & 14.82 & 34.84 & 12.32 & 18.32 & 12.11 & 32.45 & \underline{16.23} & 34.38 & 22.10 & 40.25 & 25.04 & 41.65\\
        SED$_{\textcolor{gray}{\text{\tiny \textit{CVPR2024}}}}$ & ConvNeXt-B & OVS & 21.48 & 45.15 & 93.31 & 96.66 & 20.10 & 42.30 & 5.78 & 17.52 & 14.80 & 18.90 & 10.23 & 31.50 & 9.36 & 21.62 & 17.32 & 38.12 & 24.05 & 38.97 \\
        Cat-Seg$_{\textcolor{gray}{\text{\tiny \textit{CVPR2024}}}}$ & ViT-B & OVS & 20.41 & 44.08 & \underline{94.16} & \underline{96.72} & 23.50 & 41.55 & 15.23 & 37.17 & 15.47 & 23.57 & 12.10 & 38.55 & 14.03 & 38.61 & 19.62 & \textbf{51.38} & 26.82 & 46.45 \\
        OVRS$_{\textcolor{gray}{\text{\tiny\textit{TGRS2025}}}}$ & ViT-B & OVRSIS & 21.06 & 45.48 & \textbf{94.60} & \textbf{96.87} & 25.30 & 42.32 & 15.57 & \underline{38.94} & \underline{16.22} & 24.05 & 11.90 & 38.80 & 14.66 & 38.68 & 21.42 & \underline{51.25} & 27.59 & \underline{47.05} \\
        GSNet$_{\textcolor{gray}{\text{\tiny\textit{AAAI2025}}}}$ & ViT-B & OVRSIS & \textbf{26.20} & \textbf{57.07} & 90.00 & 93.60 & \underline{26.80} & \underline{42.84} & 15.12 & 36.16 & 15.93 & \underline{24.33} & \underline{12.44} & \underline{39.62} & 14.25 & \underline{41.15} & \underline{22.22} & 42.33 & \underline{27.87} & 47.14 \\
        RSKT-Seg & ViT-B & OVRSIS & \underline{24.80} & \underline{55.94} & 93.16 & 96.37 & \textbf{28.07} & \textbf{46.58} & \textbf{20.28} & \textbf{46.71} & \textbf{17.15} & \textbf{34.78} & \textbf{13.94} & \textbf{43.47} & \textbf{17.47} & \textbf{50.84} & \textbf{25.34} & 47.18 & \textbf{30.03} & \textbf{52.73} \\

        \midrule
        SCAN$_{\textcolor{gray}{\text{\tiny \textit{CVPR2024}}}}$  &  ViT-L  &  OVS & 21.44 & 53.26 & 64.28 & 85.46 & 18.50 & 38.20 & 28.32 & \underline{52.47} & 14.32 & 26.00 & 16.22 & 32.51 & 14.23 & 34.25 & 20.18 & 36.70 & 24.69 & 44.86 \\
        SAN$_{\textcolor{gray}{\text{\tiny \textit{CVPR2023}}}}$  &  ViT-L  &  OVS & 20.54 & 49.32 & 87.22 & 92.54 & 22.33 & 44.22 & 24.72 & \textbf{56.54} & 15.01 & 26.50 & 21.35 & 37.66 & 22.49 & 50.78 & 26.42 & 42.30 & 30.01 & 49.98 \\
        SED$_{\textcolor{gray}{\text{\tiny \textit{CVPR2024}}}}$  &  ConvNeXt-L  &  OVS & 23.80 & 50.36 & 94.32 & 96.84 & 23.24 & 45.13 & 11.85 & 23.87 & 15.21 & 26.80 & 22.42 & 39.15 & 12.61 & \textbf{25.73} & 28.50 & 44.20 & 28.99 & 44.01 \\
        Cat-Seg$_{\textcolor{gray}{\text{\tiny \textit{CVPR2024}}}}$  &  ViT-L  &  OVS & 28.80 & 59.56 & \underline{94.77} & \underline{96.96} & 25.11 & 48.25 & 23.90 & 49.49 & 16.10 & 26.90 & 24.32 & 42.77 & 21.74 & 51.25 & 30.16 & 47.20 & 33.11 & 52.80 \\
        OVRS$_{\textcolor{gray}{\text{\tiny\textit{TGRS2025}}}}$  &  ViT-L  &  OVRSIS & \textbf{32.25} & 60.35 & \textbf{94.86} & \textbf{97.06} & \underline{27.98} & 49.01 & 26.39 & 50.15 & \underline{16.24} & 27.00 & 31.88 & 55.32 & \textbf{28.80} & \textbf{54.20} & 31.01 & \underline{55.30} & \underline{36.18} & 56.05 \\
        GSNet$_{\textcolor{gray}{\text{\tiny\textit{AAAI2025}}}}$  &  ViT-L  &  OVRSIS & 31.50 & \textbf{63.05} & 93.11 & 95.98 & 27.21 & \underline{49.53} & \underline{28.50} & 52.00 & 15.98 & \underline{27.10} & \underline{32.24} & \underline{56.20} & 25.10 & \underline{52.80} & \underline{32.07} & 55.21 & 35.71 & \underline{56.48} \\

        RSKT-Seg  &  ViT-L  &  OVRSIS & \underline{31.57} & \underline{61.66} & 93.96 & 96.63 & \textbf{29.62} & \textbf{51.01} & \textbf{28.57} & 51.69 & \textbf{17.36} & \textbf{35.75} & \textbf{34.01} & \textbf{57.38} & \underline{25.55} & 52.66 & \textbf{33.40} & \textbf{56.50} & \textbf{36.76} & \textbf{57.91} \\

        \bottomrule
    \end{tabular}
    \end{adjustbox}
\end{table*}

\subsection{Training Loss}
We adopt standard cross-entropy loss for supervision. Given a one-hot segmentation mask $\text{M} \in \mathbb{R}^{H \times W \times N_t}$ and the prediction $\text{O}$, the loss is defined as:
\begin{equation}
\mathcal{L}_{ce} = \text{CrossEntropyLoss}(\text{O}, \text{M})
\end{equation}
Minimizing this loss encourages the model to produce accurate pixel-level class predictions.

\section{Experiment}
\label{sec:experiment}

\subsection{Experimental Analysis of RSKT-Seg}

\subsubsection{Implementation Details of RSKT-Seg}
Our approach is implemented based on the PyTorch and Detectron2 \cite{wu2019detectron2} frameworks. The number of layers in RSKT-Upsample $N_d$ is configured as 2. When using the ViT-B/16  architecture as encoder, \(l\) is set to the values \(\{3,7\}\), and when using the ViT-L/14 architecture, \(l\) is set to \(\{7, 15\}\). We use DINO ViT-B/32 pretrained in \cite{ye2025GSNet} and \(l_d\) is set to \(\{3,7\}\). \(l_r\) for RemoteCLIP ViT-B/16 is set to \(\{3,7\}\). AdamW is utilized as the optimizer, with an initial learning rate of \(2\times10^{-4}\) and a weight decay of \(1\times10^{-4}\). The batch size is fixed at 8. All the experiments are carried out on 4 Nvidia 4090 GPUs, each having 24GB of memory. Our experimental results is that the optimal number of layers $N$ is 5 when using DLRSD as the training set, and 2 when using iSAID.

\subsubsection{Results of the Proposed Method on Different Datasets}
Table \ref{tab:01_main_table} shows that RSKT-Seg outperforms both classic OVS methods and existing OVRSIS methods across 8 OVRSIS datasets, demonstrating superior effectiveness in remote sensing open-vocabulary segmentation. Under different training datasets (DLRSD and iSAID) and backbone configurations (ViT-B and ViT-L), it consistently leads in key metrics like m-mIoU and m-mACC
RSKT-Seg exhibits strong robustness when switching training datasets, excelling in diverse scenarios such as complex land cover (Potsdam), small objects (UAVid), and challenging datasets (VDD, UDD5). Its performance scales effectively with stronger backbones, with ViT-L bringing significant gains over ViT-B. It overcomes the limitations of classic OVS methods in handling remote sensing-specific challenges.

\subsection{Ablation Study On RSKT-Seg}
Our ablation studies are conducted under the same default settings. All reported average values are obtained from inference results on the four datasets: DLRSD, iSAID, Potsdam, and Vaihingen.

\subsubsection{The Effectiveness of Different Components}  
\begin{table}[htbp]
    \centering
    \vspace{-10pt}
    \caption{\textbf{The Effectiveness of Proposed Components.}}
    \label{tab:02_main_ablation}
    \vspace{-10pt}
    \resizebox{\linewidth}{!}{
    \begin{tabular}{lccccccc}
        \toprule
        Dataset & R-I Cost Map & DINO Cost Map & RS-Transfer & RS-Fusion & m-mIoU & m-fwIoU & m-mACC \\
        \midrule
        \multirow{8}{*}{DLRSD}
        & & & & & 46.42 & 51.51 & 60.23 \\
        & \checkmark & & & & 48.06 & 53.00 & 61.72 \\
        &  & \checkmark & & & 47.47 & 54.11 & 62.16 \\
        &  &  & \checkmark & & 47.68 & 52.92 & 61.60 \\
        & \checkmark & \checkmark & & & 48.72 & 54.70 & 62.56 \\
        & \checkmark & \checkmark & \checkmark & & \underline{48.76} & \underline{54.94} & \underline{63.11} \\
        & \checkmark & \checkmark & \checkmark & \checkmark & \textbf{48.94} & \textbf{55.46} & \textbf{64.12} \\
        \midrule
        \multirow{7}{*}{iSAID}
        & & & & & 35.51 & 35.54 & 57.36 \\
        & \checkmark & & & & 38.42 & 37.96 & 58.86 \\
        &  & \checkmark & & & 37.59 & 37.66 & 58.42 \\
        &  &  & \checkmark & & 35.36 & 36.30 & 57.91 \\
        & \checkmark & \checkmark & & & 38.93 & 38.86 & 58.47 \\
        & \checkmark & \checkmark & \checkmark & & \underline{39.07} & \underline{40.31} & \underline{61.24} \\
        & \checkmark & \checkmark & \checkmark & \checkmark & \textbf{39.80} & \textbf{41.59} & \textbf{62.55} \\
        \bottomrule
    \end{tabular}
    }
    \vspace{-10pt}
\end{table}
Table \ref{tab:02_main_ablation} validates the effectiveness of each proposed component. Adding R-I Cost Map or DINO Cost Map alone improves metrics on datasets, with R-I Cost Map performing slightly better. Combining them yields further gains (e.g., DLRSD m-mIoU: 48.72 vs. 46.42). Incorporating RS-Transfer brings marginal improvements and adding RS-Fusion achieves the best results, boosting performance.

\subsubsection{The importance of introducing remote sensing knowledge}  
\begin{table}[htbp]
    \centering
    \caption{\textbf{The Effectiveness of DINO with Remote Sensing Knowledge.} "natureDINO" denotes pretrained DINO weights in \cite{caron2021emerging}, "rsDINO" denotes pretrained DINO weights on large-scale Remote Sensing Dataset in \cite{ye2025GSNet}.}
    \label{tab:03_different_dino} 
    \vspace{-10pt}
    \resizebox{1.0\linewidth}{!}{
    \begin{tabular}{cccccccc}
        \toprule
        Dataset & rotate & rsDINO & natureDINO & remoteclip & m-mIoU & m-fwIoU & m-mACC \\
        \midrule
        \multirow{7}{*}{DLRSD} 
        & & & &  & 47.17 & 52.26 & 60.48\\
        & & \checkmark & & & \textbf{48.22} & \textbf{54.86} & \textbf{62.41}  \\
        & & & \checkmark & & 46.46 & 53.87 & 62.19 \\
        & \checkmark & \checkmark & & & \textbf{49.47} & \textbf{55.45} & \textbf{62.81} \\
        & \checkmark & & \checkmark &  & 46.90 & 54.96 & 62.40 \\
        & \checkmark & \checkmark &  &  \checkmark &  \textbf{49.37} &  \textbf{55.29} &  \textbf{62.74}\\
        & \checkmark & & \checkmark & \checkmark & 47.17 & 54.11 & 61.70 \\
        \bottomrule
    \end{tabular}
    }
    \vspace{-10pt}
\end{table}

The \cref{tab:03_different_dino} compares the performance of "rsDINO" and "natureDINO" on the DLRSD dataset.
When the "DINO" component is added to the model, it improves the model's performance significantly.
Evidently, the "rsDINO" pre-trained on remote sensing data is more effective in enhancing the model's performance for the DLRSD dataset.
\begin{figure}[t]
    \centering
    \includegraphics[width=\linewidth]{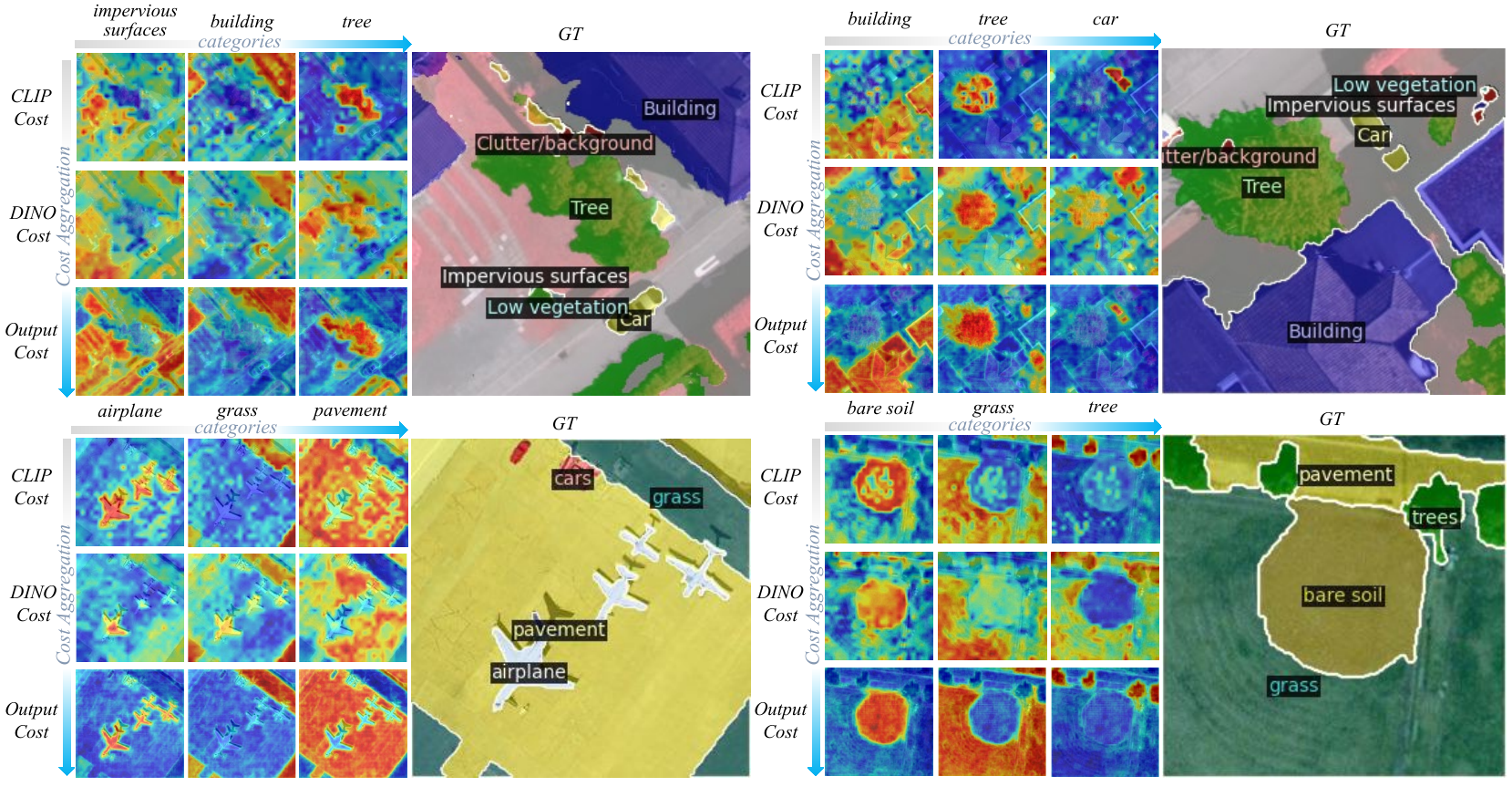}
    \vspace{-10pt}
    \caption{\textbf{Comparasion of different cost map and effectiveness of efficient cost map aggregation} (vertical) on different classes(horizontal). }
    \label{fig:fig_04_cost_vis}
\end{figure}

\subsubsection{Ablation of Different cost map fusion strategy}  
\begin{table}[ht]
    \centering
    \small
    \vspace{-12pt}
    \caption{\textbf{The Influence of Different Cost Map Fusion Strategies $Fusion$.}}
    \label{tab:04_different_fusion}
    \vspace{-10pt}
    \resizebox{0.80\linewidth}{!}{
    \begin{tabular}{lccc}
        \toprule
        $Fusion$ & m-mIoU & m-fwIoU & m-mACC \\
        \midrule
        mean & 47.12 & 51.99 & \textbf{66.56} \\
        cat & \textbf{48.74} & \textbf{52.50} & 66.24 \\
        separate & 47.83 & 51.89 & 65.17 \\
        \bottomrule
    \end{tabular}
    }
\end{table}

The \cref{tab:04_different_fusion} shows evaluation metrics for different cost map fusion strategies. The values of m-mIoU, m-fwIoU, and m-mACC for “mean”, “cat”, and “separate” strategies vary slightly. This indicates that these strategies have similar effects on model performance. 

\subsubsection{The impacts of different layers of cost map aggregation}  
\begin{table}[ht]
    \centering
    \small
    \caption{\textbf{The Influence of Different Layers of the Cost Aggregation Module.} We test on 8 datasets in this experiment.}
    \label{tab:05_different_layer}
    \resizebox{0.98\linewidth}{!}{
    \begin{tabular}{lcccccc}
        \toprule
        Dataset/ $N$ & 1 & 2 & 3 & 4 & 5 & 6 \\
        \midrule
        DLRSD m-mIoU & 46.21 & 46.87 & \underline{49.25} & 48.84 & \textbf{51.58} & 48.20 \\
        DLRSD m-mACC & 61.35 & 62.36 & 64.06 & 63.11 & \textbf{65.41} & \underline{64.11} \\
        \midrule
        iSAID m-mIoU & 38.13 & \textbf{39.09} & \underline{38.93} & 37.22 & 37.84 & 37.34 \\
        iSAID m-mACC & 61.32 & \underline{62.21} & \textbf{62.47} & 60.17 & 60.29 & 60.24 \\
        \bottomrule
    \end{tabular}
    }
\end{table}

This subsection analyzes the impacts of different layers in the cost map aggregation module using data from \cref{tab:05_different_layer} on DLRSD and iSAID datasets, with metrics like m-mIoU, m-fwIoU, and m-mACC. Based on the experimental results, we selected N=5 for the DLRSD dataset and N=2 for the iSAID dataset as the optimal configurations for the Cost Aggregation Module

\subsubsection{The impacts of different finetuning strategies}  
\begin{table}[htbp]
    \centering
    \vspace{-10pt}
    \caption{\textbf{The Ablation Study of Different Fine-tuning Strategies.} The trainable CLIP and the frozen DINO demonstrate better performance. }
    \label{tab:06_fintune_method}
     \vspace{-10pt}
    \resizebox{1.0\linewidth}{!}{
    \begin{tabular}{cllcccc}
        \toprule
        Dataset & CLIP & DINO & m-mIoU & m-fwIoU & m-mACC \\
        \midrule
        \multirow{4}{*}{iSAID}
        & frozen & frozen & 28.80 & 29.70 & 45.69 \\
        & frozen & attention & 27.42 & 29.31 & 46.79 \\
        & attention & frozen & \textbf{39.07} & \textbf{40.31} & \textbf{61.24} \\
        & attention & attention & 37.96 & 38.98 & 61.21 \\
        \bottomrule
    \end{tabular}
    }
\end{table}

This subsection analyzes the impacts of different finetuning strategies using data from \cref{tab:06_fintune_method} on the iSAID dataset. The key components are CLIP and DINO, each with frozen and attention states.
when CLIP is in attention and DINO frozen (highlighted), the model achieves the best performance among all combinations.

\subsubsection{The analysis of inference speed}  
\begin{table}[htbp]
\centering
\vspace{-10pt}
\caption{\textbf{The Comparative Experiment on Inference Speed and FPS.} }
\label{tab:07_inference_speed}
\vspace{-10pt}
\resizebox{1.0\linewidth}{!}{
\begin{tabular}{lllllll}
\toprule
Method & DLRSD & iSAID & Potsdam & Vaihingen & Mean (ms) $\downarrow$ & FPS  $\uparrow$ \\
\midrule
SCAN & 148.27 & 143.24 & 143.36 & 148.46 & 145.83 & 6.86 \\
SAN & 54.15 & 49.30 & 50.64 & 61.06 & 53.79 & 18.59 \\
SED & 61.91 & 63.13 & 58.50 & 58.35 & 60.47 & 16.54 \\
Cat-Seg & 125.08 & 118.02 & 108.14 & 108.27 & 114.88 & 8.70 \\
OVRS & 301.60 & 292.33 & 273.84 & 273.72 & 285.37 & 3.50 \\

RSKT-Seg & \textbf{69.95} & \textbf{64.90} & \textbf{62.88} & \textbf{62.70} & \textbf{65.11} & \textbf{15.36} \\
\bottomrule
\end{tabular}
}
\vspace{-7pt}
\end{table}

This table presents the inference speed (in ms) and FPS across multiple datasets for different methods. Among them, RSKT-Seg stands out for its significantly improved inference speed compared to Cat-Seg and OVRS. RSKT-Seg only needs 65.11 ms on average. This shows that RSKT-Seg can achieve faster inference, which is beneficial for real-time applications.

\begin{figure}[t]
    \centering
    \includegraphics[width=\linewidth]{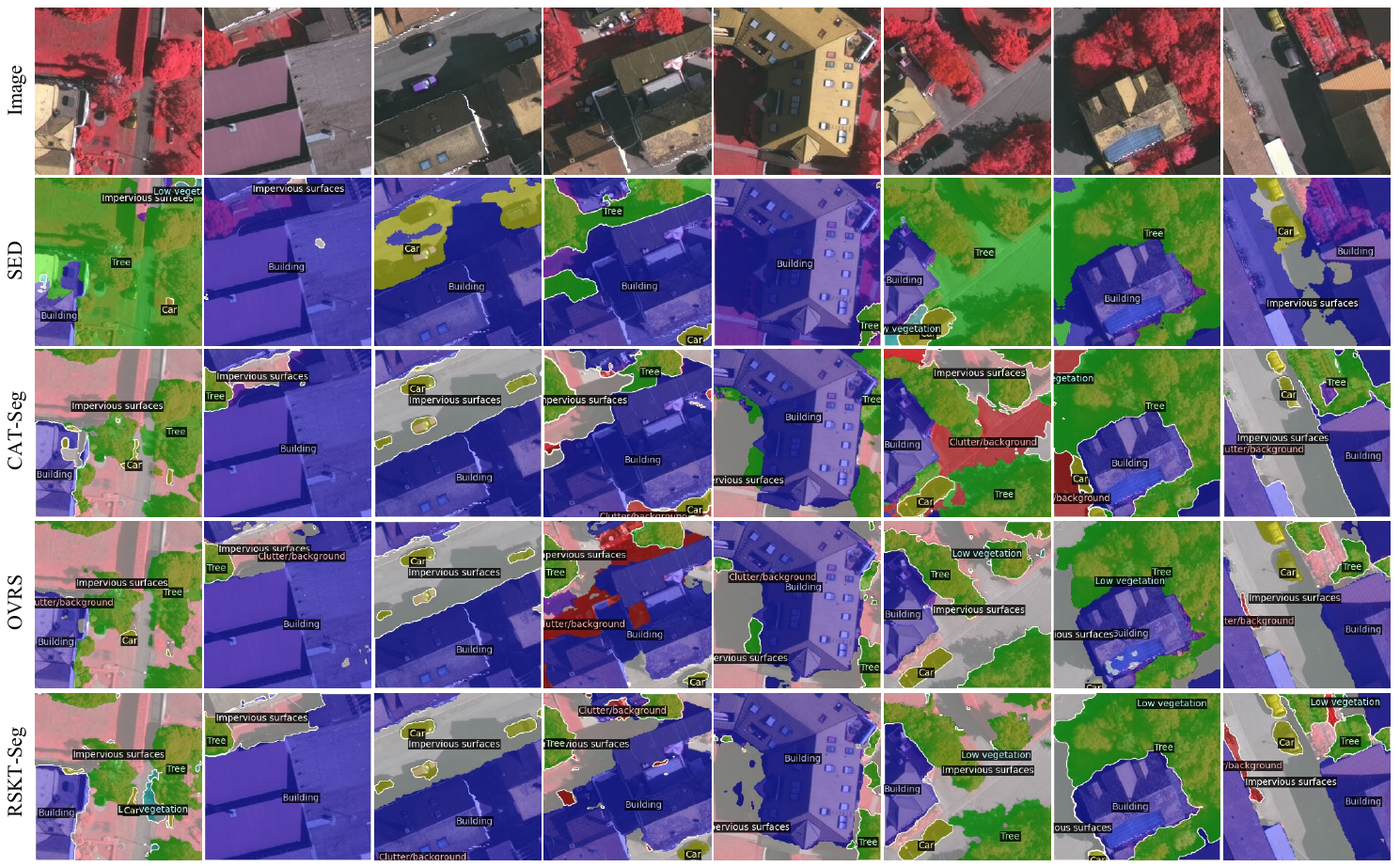}
    \vspace{-10pt}
    \caption{\textbf{The comparison of the segmentation visualization of different models.} The model (ViT-L) trained on DLRSD is used to perform inference on Vaihingen.}
    \label{fig:fig_04_segmentation}
    \vspace{-10pt}
\end{figure}

\section{Visualization}
\label{sss}
\subsection{Cost map visualization on diffetent categories}
As seen in \cref{fig:fig_04_cost_vis}, cost maps from CLIP Cost and DINO Cost show how models assign costs to classes like impervious surfaces, trees, and buildings. These heat-like visualizations reveal areas of high or low classification confidence. RS-Fusion, depicted vertically in \cref{fig:fig_04_cost_vis} across different classes, refines these maps, enhancing segmentation accuracy. More Cost visualization figures are shown in appendix E. 

\subsection{Comparison of visualization}
As depicted in \cref{fig:fig_04_segmentation}, a comparison is made among the segmentation visualizations of multiple models. These models were trained on the DLRSD dataset and then tested on the Vaihingen dataset. Each row corresponds to a different model, while the columns represent various image samples. RSKT-Seg demonstrates notable effectiveness in this comparison. Compared to other models, RSKT-Seg shows a relatively lower rate of misclassification and a more precise identification of object boundaries.
\section{Conclusion}
\label{sec:conclusion}
In this paper, we address OVRSIS challenges via \textbf{OVRSISBench}, a unified benchmark. Evaluations show classic OVS methods perform poorly in remote sensing, and existing OVRSIS methods lack RS-specific modeling. Thus, we propose the \textbf{RSKT-Seg} that integrates RS-CMA, RS-Fusion, and RS-Transfer, outperforming baselines with 2× faster inference, advancing OVRSIS research. 
In the appendix H, we carefully analyzed the gap of open-vocabulary segmentation from natural images to remote-sensing images, as well as the \textbf{limitations} of our model, and our \textbf{future work}.

\section{Acknowledgments}
This work was supported in part by the Natural Science Foundation of China under Grant 62306241, and in part by grants from the Innovation Foundation for Doctor Dissertation of Northwestern Polytechnical University (No.CX2025109).

\appendix

\section{A: More related works}
\subsection{Open-Vocabulary Segmentation}
Semantic segmentation models traditionally rely on predefined datasets, limiting their scalability and adaptability to novel scenarios. The advent of pretrained Vision-Language Models (VLMs) has introduced a new paradigm for segmentation. However, these models, pretrained at the image level, lack the fine-grained pixel-text alignment required for effective segmentation, posing a significant challenge for direct application.
To address this issue, researchers have explored two main strategies: two-stage and single-stage approaches. Two-stage methods \cite{liang2023open, xu2022simple, zhou2022extract} utilize external mask generators to refine segmentation predictions. However, the limited accuracy of these generated masks reduces the overall effectiveness and generalization of such approaches.
In contrast, single-stage models integrate mask generation directly within the segmentation process. FC-CLIP \cite{yu2023convolutions} and its enhanced version \cite{jiao2025collaborative} incorporate trainable mask predictors within the CLIP framework, improving segmentation accuracy. More advanced techniques, such as Cat-Seg \cite{cho2024cat} and SED \cite{xie2024sed}, further refine segmentation performance by leveraging similarity matrices as pseudo-masks.
Recent advancements have introduced novel mechanisms to enhance segmentation effectiveness. Side adapters, proposed in \cite{xu2023side}, facilitate efficient feature fusion, while frequency-domain modules \cite{xu2024generalization} improve model generalization. Additionally, \cite{shan2024open} presents an adaptive integration of SAM and CLIP outputs, optimizing segmentation efficiency. These ongoing innovations continue to refine vision-language alignment, advancing both accuracy and adaptability in open-vocabulary segmentation.

\section{B: The analysis of computational efficiency}

\begin{table}[htbp]
    \centering
    \vspace{-10pt}
    \caption{\textbf{The Comparative Experiment on the efficiency and computational complexity of the model.} }
    \label{tab:time_GFLOPs} 
    \vspace{-10pt}
    \resizebox{1.0\linewidth}{!}{
    \begin{tabular}{lcccc}
        \hline
        Method & train time(ms/iteration) & total param.(M) & trainable param(M) \\
        \hline
        SED & 8.60 & 180.76 & 89.59 \\
        Cat-Seg & 9.60 & \textbf{154.29} & 127.55 \\
        OVRS & 18.53 & 154.32 & 127.57 \\
        
        RSKT-Seg & \textbf{7.96} & 334.62 & \textbf{59.89} \\
        \hline
    \end{tabular} }
    \vspace{-10pt}
\end{table}

The presented \cref{tab:time_GFLOPs} compares the computational efficiency of RSKT-Seg with multiple models, taking into account training time per iteration and parameter complexity. Among them, RSKT-Seg clearly demonstrates its superiority. It achieves the shortest training time of merely 7.96 ms/iteration. Although it has a relatively large total parameter count of 334.62 M, the significantly fewer trainable parameters, only 59.89 M, substantially enhance its training efficiency. This allows for rapid convergence during training. In contrast, Cat-Seg, having the fewest total parameters (154.29 M), lags in training speed with 9.60 ms/iteration. OVRS is the slowest, with a training time of 18.53 ms/iteration. SED remains at an average level. Overall, RSKTSeg's remarkable efficiency makes it an excellent choice, especially in resource-constrained scenarios.
Furthermore, for the effectiveness analysis of the dimensional reduction strategy in efficient cost-map aggregation, please refer to the Appendix. \textcolor[rgb]{0.4, 0, 0.4}{C}.

\section{B: Computational Complexity Analysis}
\label{app:CCA}
\subsection{Analysis for Spatial Dimension Reduction}
The original spatial size of the features is $H_f \times W_f$. After the operation of the 2D convolutional layer $\mathcal{R}_s$, the spatial size becomes $\frac{H_f}{r_1} \times \frac{W_f}{r_1}$, where $r_1$ is the reduction ratio in the spatial dimension.

In the cross-attention mechanism of the Transformer, the computational complexity of calculating the attention scores is typically proportional to the square of the sequence length. Before the spatial dimension reduction, assume the sequence length is $L_1 = H_f \times W_f$. After the reduction, the sequence length becomes $L_2=\frac{H_f \times W_f}{r_1^2}$.

The computational complexity of the cross-attention mechanism is $O(L^2d)$, where $L$ is the sequence length and $d$ is the feature dimension. The computational complexity before reduction is $C_1 = O((H_f \times W_f)^2d_c)$, and after reduction, it is $C_2 = O((\frac{H_f \times W_f}{r_1^2})^2d_c)$.

The reduction ratio of the computational complexity is:
\begin{equation}
    \frac{C_2}{C_1}=\frac{O((\frac{H_f \times W_f}{r_1^2})^2d_c)}{O((H_f \times W_f)^2d_c)}=\frac{1}{r_1^4}
\end{equation}

\subsection{Analysis for Class Dimension Reduction}
In the RS Class Fusion Transformer, an average pooling layer $\mathcal{R}_{c}$ is used to reduce the class dimension of the features. 

In the self-attention mechanism of the Transformer, the computational complexity of calculating the attention scores is also proportional to the square of the sequence length. Before the class dimension reduction, assume the sequence length is $L_3 = H_f \times W_f$. After the reduction, the sequence length becomes $L_4=\frac{H_f \times W_f}{r_2^2}$.

The computational complexity before reduction is $C_3 = O((H_f \times W_f)^2d_c)$, and after reduction, it is $C_4 = O((\frac{H_f \times W_f}{r_2^2})^2d_c)$.

The reduction ratio of the computational complexity is:
\begin{equation}
\frac{C_4}{C_3}=\frac{O((\frac{H_f \times W_f}{r_2^2})^2d_c)}{O((H_f \times W_f)^2d_c)}=\frac{1}{r_2^4}
\end{equation}

\begin{figure*}[ht]
    \centering
    \includegraphics[width=\linewidth]{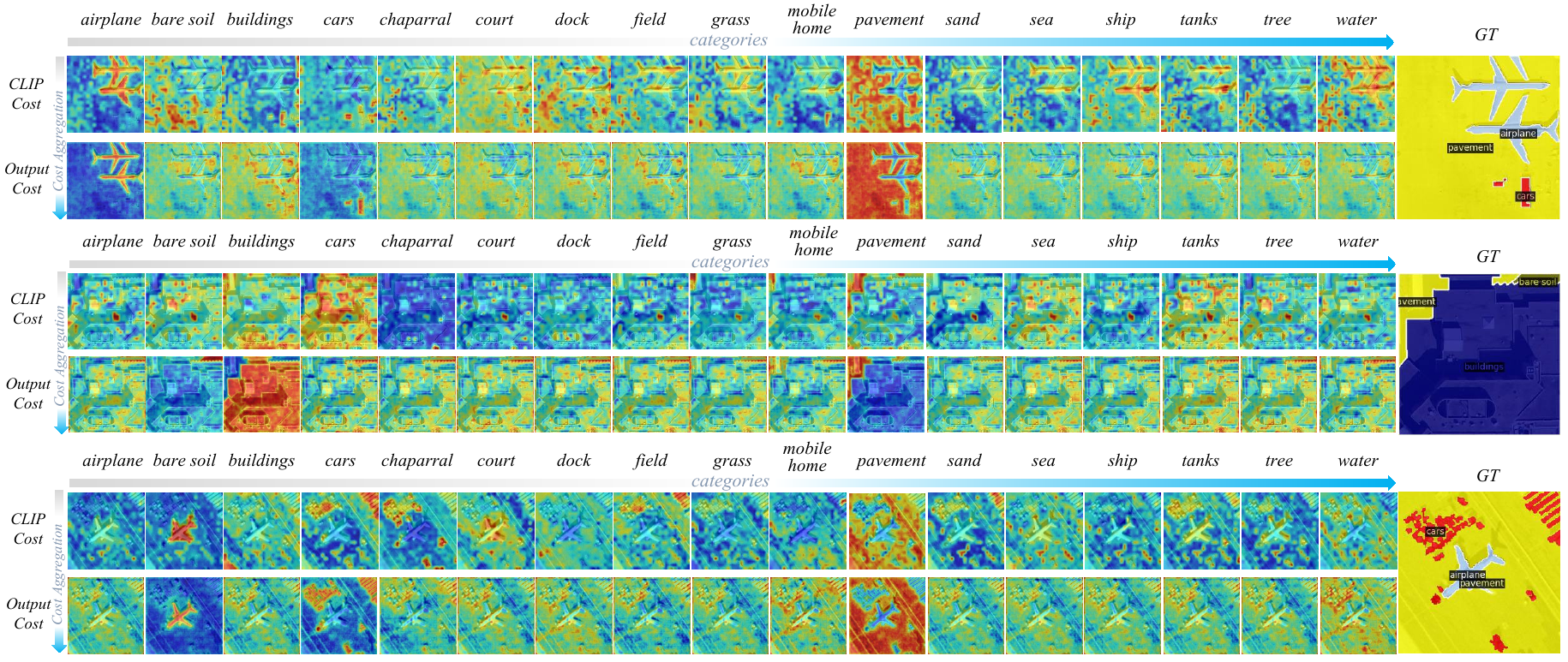}
    \caption{Comparing the Effectiveness of Cost Maps and Cost Aggregation on Different Categories using the DLRSD Dataset.}
    \label{fig:fig_07_app_cost_vis}
\end{figure*}

\begin{figure*}[ht]
    \centering
    \includegraphics[width=\linewidth]{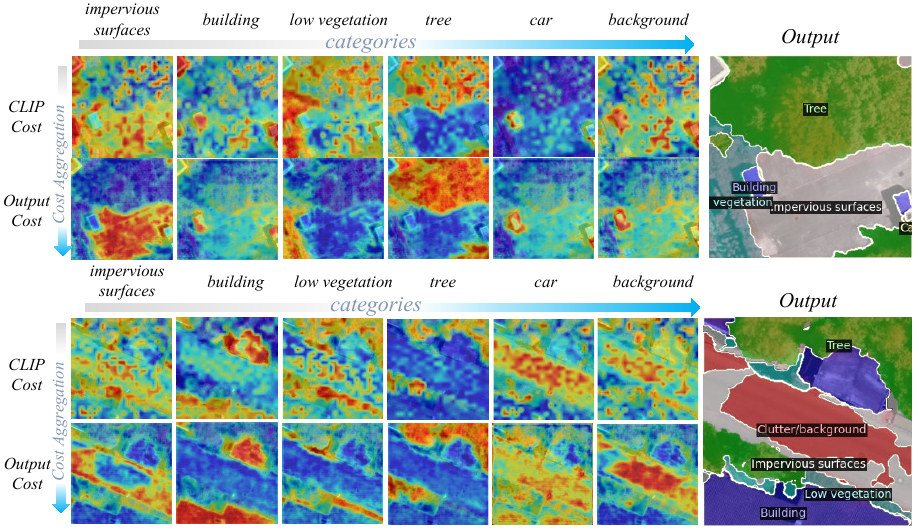}
    \caption{Comparing the Effectiveness of Cost Maps and Cost Aggregation on Different Categories using the Vaihingen Dataset.}
    \label{fig:fig_08_app_cost_vis}
\end{figure*}

\section{C: Datasets introduction}
\label{sec:app_dataset_introduction}
\subsection{DLRSD} 
It is a large-scale remote sensing image dataset, consisting of a total of 7002 images across 17 categories. In the experimental configuration, 5601 images are designated for training, and the remaining 1401 images are used for validation.
\subsection{iSAID} 
It contains 24,439 images across 15 categories. In the experimental setup, 18,076 images are used for training, and the remaining 6,363 images are employed to validate the model's performance.
\subsection{Potsdam} 
The Potsdam dataset is a large-scale and extensively used dataset in the realm of remote sensing images. With 20,102 images spread across 6 categories, it has been a staple for numerous research projects and model evaluations. 
\subsection{Vaihingen} 
The Vaihingen dataset is a commonly used remote sensing image dataset in the academic and research community. Comprising 2254 images across 6 categories, it provides a valuable resource for studying and developing remote sensing applications.

\subsection{UAVid} 
UAVid is a high-resolution dataset captured from unmanned aerial vehicles, designed for urban scene understanding. It consists of 300 images annotated with 8 semantic categories. In our experiments, 200 images are used for training and 100 for validation.

\subsection{UDD5} 
The UDD5 dataset contains urban driving scenes collected from a bird's-eye view, with pixel-wise annotations across 5 semantic categories. It includes 4,198 images in total, with 3,149 images used for training and 1,049 for validation.

\subsection{LoveDA} 
LoveDA is a large-scale remote sensing dataset with diverse land cover types, collected from rural and urban scenes. It includes 18,000 images annotated with 7 classes. For training, 14,168 images are used, while the remaining 3,832 are reserved for validation.

\subsection{VDD} 
VDD is a recently introduced dataset focused on visual domain adaptation for remote sensing tasks. It consists of 7,992 images labeled across 7 categories. In our setup, 5,994 images are used for training, and 1,998 images are used for validation.

\section{D: Evaluation metric}
\label{sec:app_evaluation_metric}
Consistent with existing OVRSIS methods, we use the mean Intersection over Union (mIoU), Frequency Weighted Intersection over Union (fwIoU), and Mean Accuracy (mACC) as evaluation metrics.
The mIoU is computed as the average IoU across all classes, and is expressed as Equation \cref{eq:mIoU}:
\begin{equation}
mIoU=\frac{1}{n}\sum_{i = 1}^{n}\frac{TP_{i}}{TP_{i}+FP_{i}+FN_{i}}
\label{eq:mIoU}
\end{equation}
where \(n\) is the number of classes, \(TP_{i}\) is the number of true positives for class \(i\), \(FP_{i}\) is the number of false positives for class \(i\), and \(FN_{i}\) is the number of false negatives for class \(i\).
The fwIoU metric, which accounts for the frequency of each class and offers a weighted measure, is calculated as Equation \cref{eq:fwIoU}:
\begin{equation}
\small
fwIoU=\sum_{i = 1}^{n}\frac{TP_{i}+FN_{i}}{\sum_{j = 1}^{n}(TP_{j}+FN_{j})}\times\frac{TP_{i}}{TP_{i}+FP_{i}+FN_{i}}
\label{eq:fwIoU}
\end{equation}
The mACC, which evaluates the per-class accuracy, is given by Equation \cref{eq:mACC}:
\begin{equation}
mACC=\frac{1}{n}\sum_{i = 1}^{n}\frac{TP_{i}}{TP_{i}+FN_{i}}
\label{eq:mACC}
\end{equation}

\section{E: More cost map visualization}
\label{sec:more_cost_map}
The \cref{fig:fig_07_app_cost_vis} and \cref{fig:fig_08_app_cost_vis} offer insightful comparisons of cost maps and cost aggregation across different object categories. \cref{fig:fig_07_app_cost_vis} focuses on the DLRSD dataset, displaying cost maps from CLIP Cost and Output Cost Aggregation for a wide range of categories such as airplane, buildings, cars, etc. Similarly, \cref{fig:fig_08_app_cost_vis} examines the Vaihingen dataset, presenting cost maps for categories like impervious surfaces, building, and tree. By comparing CLIP Cost and Output Cost Aggregation, it becomes evident how cost aggregation refines the initial cost maps, potentially improving the accuracy of object segmentation. Moreover, cost maps become more discriminative among different categories after efficient aggregation. 

\begin{table*}[h]
\centering
\caption{\textbf{Comparision of per-class mIoU on Vaihingen dataset.} We use the model trained on DLRSD dataset and perform inference on the Vaihingen dataset.}
\label{tab:per_class_vaihingen}
\vspace{-10pt}
\resizebox{1.0\linewidth}{!}{
\begin{tabular}{lcccccccc}
\toprule
Method & Backbone & Impervious surfaces & Building & Low vegetation & Tree & Car & Clutter/background & mIoU \\
\midrule
SCAN$_{\textcolor{gray}{\text{\tiny \textit{CVPR2024}}}}$ & ViT-B & 5.60 & 9.64 & 1.15 & 7.28 & 4.04 & 4.55 & 5.38 \\
SAN$_{\textcolor{gray}{\text{\tiny \textit{CVPR2023}}}}$ & ViT-B & 33.26 & 56.36 & 7.13 & 43.29 & 24.21 & 27.23 & 31.92 \\
SED$_{\textcolor{gray}{\text{\tiny \textit{CVPR2024}}}}$ & ConvNeXt-B & 30.50 & 52.34 & 6.37 & 40.00 & 22.13 & 25.08 & 29.40 \\
Cat-Seg$_{\textcolor{gray}{\text{\tiny \textit{CVPR2024}}}}$ & ViT-B & 33.46 & 57.19 & 7.32 & 43.77 & 24.44 & 27.74 & 32.32 \\
OVRS$_{\textcolor{gray}{\text{\tiny \textit{CVPR2024}}}}$ & ViT-B & 35.06 & 59.32 & 7.72 & 45.45 & 25.81 & 28.89 & 33.71 \\
RSKT-Seg & ViT-B & 35.27 & 59.57 & 7.54 & 46.71 & 26.57 & 30.34 & 34.33 \\
\bottomrule
\end{tabular}
}
\end{table*}

\section{F: Comparison of RSKT-Seg with Different Models at Each Category}
The performance of RSKT-Seg is comprehensively evaluated by comparing it with several other models at the granularity of each category.  \cref{tab:per_class_vaihingen} presents the per-class mIoU (mean Intersection over Union) values on the Vaihingen datasets, where all models are trained on the DLRSD dataset.

\begin{figure}[bp]
    \centering
    \includegraphics[width=0.9\linewidth]{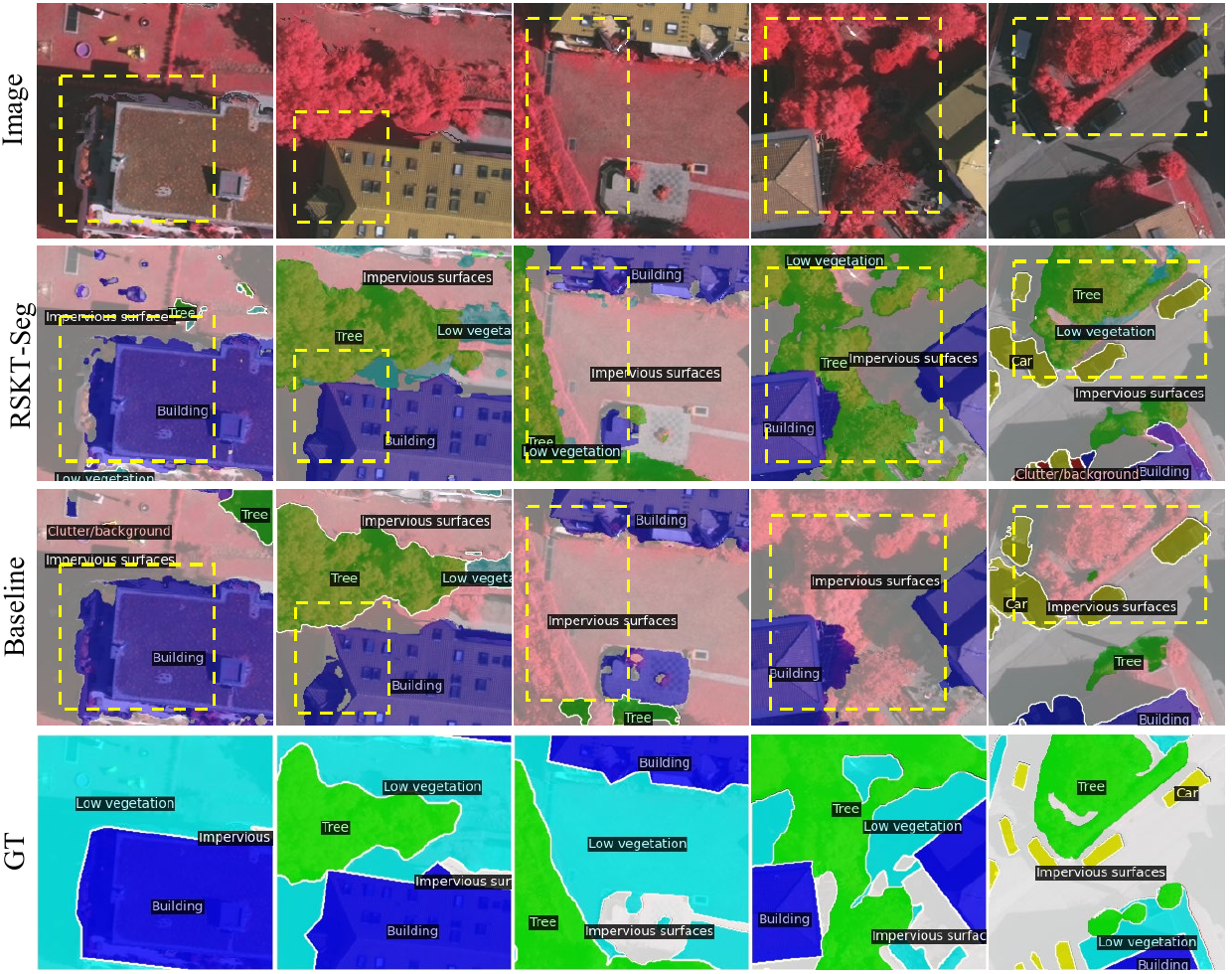}
    \caption{\textbf{Comparision of visualization of RSKT-Seg and baseline}}
    \label{fig:fig_05_RSKT_baseline}
\end{figure}

\begin{figure}[bp]
    \centering
    \includegraphics[width=0.9\linewidth]{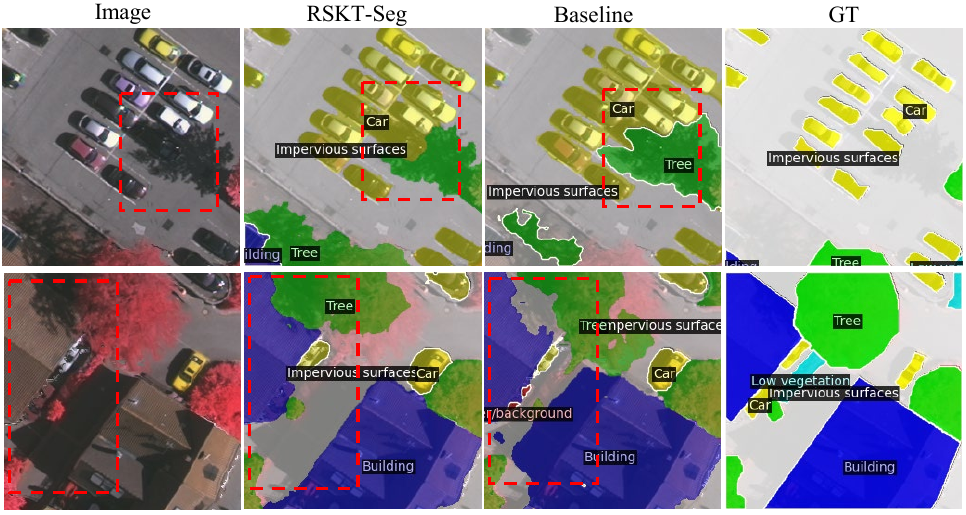}
    \vspace{-10pt}
    \caption{\textbf{Failure case}. The presence of shadows causes misclassification.}
    \label{fig:fig_06_failure_case}
\end{figure}

\begin{figure}[bp]
    \centering
    \includegraphics[width=0.9\linewidth]{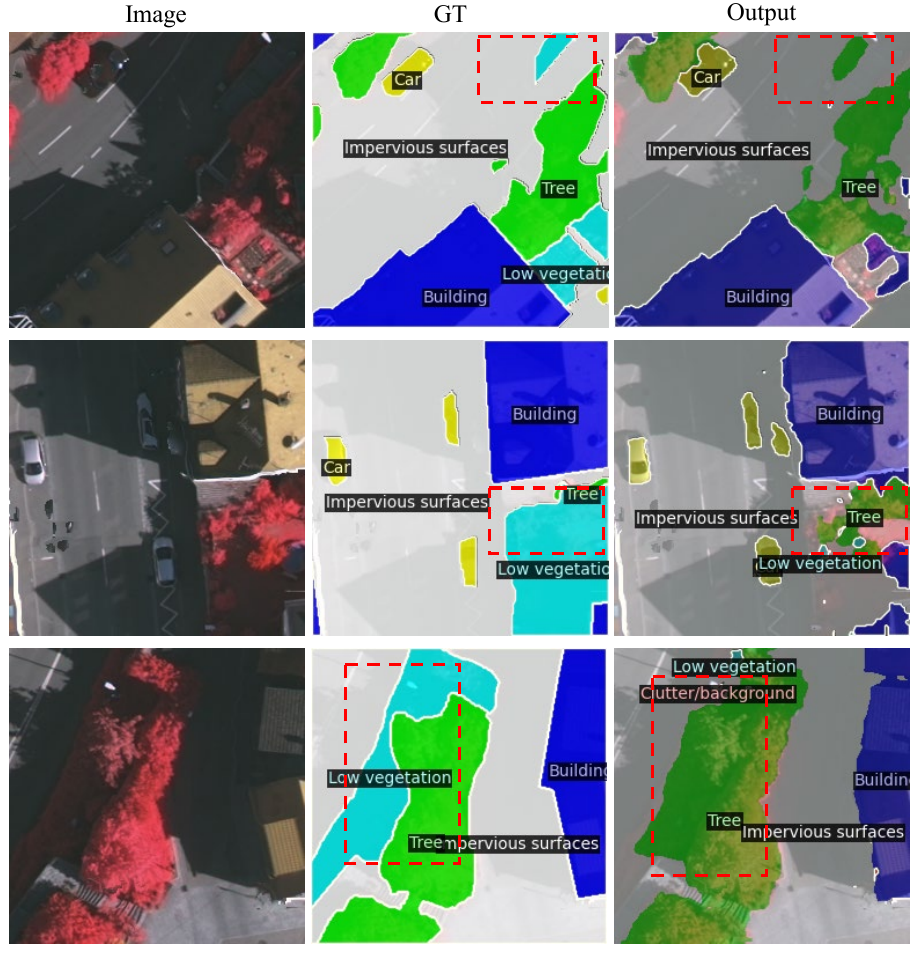}
    \caption{\textbf{limitation}. The model can not distinguish the height of remote sensing image (Low vegetation and tree).}
    \label{fig:fig_06_high_low}
\end{figure}

Regarding the Vaihingen dataset (\cref{tab:per_class_vaihingen}), RSKT-Seg also shows excellent performance. In the “Impervious surfaces” category, it attains an mIoU of 35.27, surpassing SCAN, SED, and being slightly ahead of SAN, Cat-Seg, and OVRS. In the “Tree” category, RSKT-Seg's mIoU of 46.71 is higher than the values of other models. The overall mIoU of RSKT-Seg for the Vaihingen dataset is 34.33, demonstrating its effectiveness in segmenting different surface types in this dataset. Overall, these comparisons highlight the strong adaptability and high-accuracy segmentation ability of RSKT-Seg across different datasets and object categories.

\section{H: The Effectiveness of Remote Sensing Knowledge Transfer (RSKT)}
\label{sec:app_RSKT_baseline}
The \cref{fig:fig_05_RSKT_baseline} compares RSKT-Seg and the baseline model. RSKT-Seg excels in object segmentation precision, more accurately outlining building boundaries, clearly distinguishing vegetation types, and consistently delineating impervious surfaces compared to the baseline. It also demonstrates a superior overall scene understanding, with more coherent results aligning better with the actual scene. In contrast, the baseline shows misclassification, less-defined edges, and disjointed segmentation. This visual comparison strongly indicates that RSKT-Seg outperforms the baseline, highlighting its effectiveness in remote sensing domain adaptation, which is crucial for applications like urban planning and environmental monitoring.

\section{I: Limitation and Future Work}
\label{app:limitation_future}
The performance of the model is not without \textbf{limitations}, as illustrated in \cref{fig:fig_06_failure_case} and \cref{fig:fig_06_high_low}.
\cref{fig:fig_06_failure_case} showcases a failure case where the model is misled by elements such as shadows. In the segmentation results of both RSKT-Seg and the baseline, we can observe that the presence of shadows causes misclassification. For example, in the red-boxed areas, the model incorrectly segments regions affected by shadows, failing to accurately distinguish between impervious surfaces, trees, and cars. The shadowed areas interfere with the model's ability to correctly identify object boundaries and categories, leading to sub-optimal segmentation outcomes.

\cref{fig:fig_06_high_low} further highlights another limitation of the model. It demonstrates that the model struggles to differentiate between low vegetation and trees based on height. In the marked regions of the output compared to the ground truth (GT), there are evident misclassifications. The model confuses low vegetation with trees, which is a significant shortcoming. This inability to accurately discern between these two categories based on height-related features is an area that requires further improvement.

In \textbf{future work}, we will consider introducing the depth modality to enhance the sense of distance, enabling the model to identify differences highly relevant to height and distinguish the impact of shadows on the segmentation results.

\section{J: Detailed Framework}

\begin{figure*}[t]
    \centering
    \includegraphics[width=0.92\linewidth]{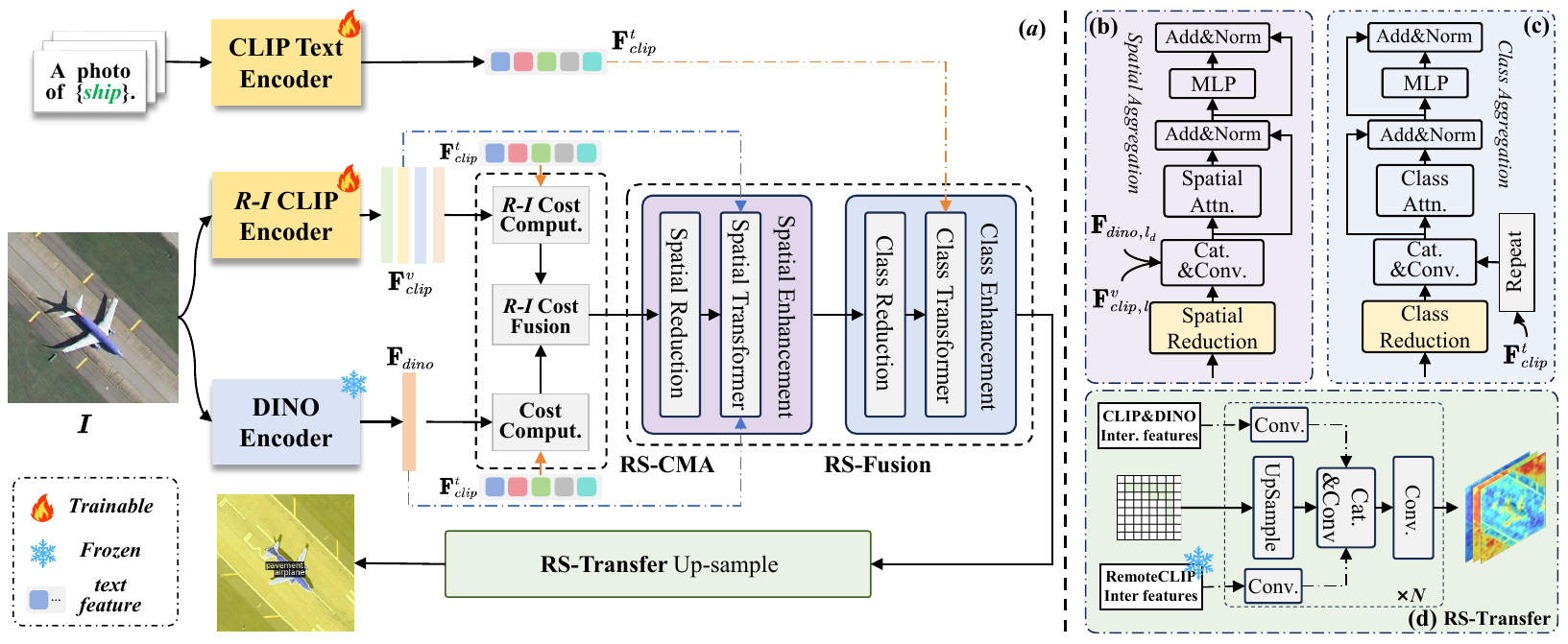}
    \vspace{-8pt}
    \caption{\textbf{ The overall framework of RSKT-Seg} includes: (a) the overall procedure of RSKT-Seg; (b) the workflow of the RS Spatial Fusion Transformer; (c) the operation flow of the RS Class Fusion Transformer; (d) the framework principle of the Remote Sensing Knowledge Transfer Upsample.}
    \vspace{-8pt}
    \label{fig:fig_app_02}
\end{figure*}

\newpage

\newpage
\bibliography{aaai2026}

@inproceedings{zheng2021rethinking,
  title={Rethinking semantic segmentation from a sequence-to-sequence perspective with transformers},
  author={Zheng, Sixiao and Lu, Jiachen and Zhao, Hengshuang and Zhu, Xiatian and Luo, Zekun and Wang, Yabiao and Fu, Yanwei and Feng, Jianfeng and Xiang, Tao and Torr, Philip HS and others},
  booktitle={Proceedings of the IEEE/CVF Conference on Computer Vision and Pattern Recognition},
  pages={6881--6890},
  year={2021}
}

@article{kotaridis2021remote,
  title={Remote sensing image segmentation advances: A meta-analysis},
  author={Kotaridis, Ioannis and Lazaridou, Maria},
  journal={ISPRS Journal of Photogrammetry and Remote Sensing},
  volume={173},
  pages={309--322},
  year={2021},
  publisher={Elsevier}
}

@article{diakogiannis2020resunet,
  title={ResUNet-a: A deep learning framework for semantic segmentation of remotely sensed data},
  author={Diakogiannis, Foivos I and Waldner, Fran{\c{c}}ois and Caccetta, Peter and Wu, Chen},
  journal={ISPRS Journal of Photogrammetry and Remote Sensing},
  volume={162},
  pages={94--114},
  year={2020},
  publisher={Elsevier}
}

@article{cao2024open,
  title={Open-vocabulary remote sensing image semantic segmentation},
  author={Cao, Qinglong and Chen, Yuntian and Ma, Chao and Yang, Xiaokang},
  journal={arXiv preprint arXiv:2409.07683},
  year={2024}
}

@inproceedings{ye2025GSNet,
  title={Towards Open-Vocabulary Remote Sensing Image Semantic Segmentation},
  author={Ye, Chengyang and Zhuge, Yunzhi and Zhang, Pingping},
  booktitle={Proceedings of the AAAI Conference on Artificial Intelligence},
  year={2025}
}

@article{li2025fgaseg,
  title={FGAseg: Fine-Grained Pixel-Text Alignment for Open-Vocabulary Semantic Segmentation},
  author={Li, Bingyu and Zhang, Da and Zhao, Zhiyuan and Gao, Junyu and Li, Xuelong},
  journal={arXiv preprint arXiv:2501.00877},
  year={2025}
}

@inproceedings{caron2021emerging,
  title={Emerging Properties in Self-Supervised Vision Transformers},
  author={Caron, Mathilde and Touvron, Hugo and Misra, Ishan and J\'egou, Herv\'e  and Mairal, Julien and Bojanowski, Piotr and Joulin, Armand},
  booktitle={Proceedings of the International Conference on Computer Vision (ICCV)},
  year={2021}
}

@inproceedings{cho2024cat,
  author    = {Cho, Seunghyun and Shin, Hyunjung and Hong, Seunghoon and Arnab, Anurag and Seo, Paul H. and Kim, Seon Joo},
  title     = {Cat-seg: Cost aggregation for open-vocabulary semantic segmentation},
  booktitle = {Proceedings of the IEEE/CVF Conference on Computer Vision and Pattern Recognition},
  pages     = {4113--4123},
  year      = {2024}
}

@article{LYU2020108,
	author = "Ye Lyu and George Vosselman and Gui-Song Xia and Alper Yilmaz and Michael Ying Yang",
	title = "UAVid: A semantic segmentation dataset for UAV imagery",
	journal = "ISPRS Journal of Photogrammetry and Remote Sensing",
	volume = "165",
	pages = "108 - 119",
	year = "2020",
	issn = "0924-2716",
	doi = "https://doi.org/10.1016/j.isprsjprs.2020.05.009",
	url = "http://www.sciencedirect.com/science/article/pii/S0924271620301295",
}

@inproceedings{chen2018large,
  title={Large-scale structure from motion with semantic constraints of aerial images},
  author={Chen, Yu and Wang, Yao and Lu, Peng and Chen, Yisong and Wang, Guoping},
  booktitle={Chinese Conference on Pattern Recognition and Computer Vision (PRCV)},
  pages={347--359},
  year={2018},
  organization={Springer}
}

@inproceedings{NEURIPS_DATASETS_AND_BENCHMARKS2021_4e732ced,
         author = {Wang, Junjue and Zheng, Zhuo and Ma, Ailong and Lu, Xiaoyan and Zhong, Yanfei},
         booktitle = {Proceedings of the Neural Information Processing Systems Track on Datasets and Benchmarks},
         editor = {J. Vanschoren and S. Yeung},
         pages = {},
         publisher = {Curran Associates, Inc.},
         title = {LoveDA: A Remote Sensing Land-Cover Dataset for Domain Adaptive Semantic Segmentation},
         url = {https://datasets-benchmarks-proceedings.neurips.cc/paper_files/paper/2021/file/4e732ced3463d06de0ca9a15b6153677-Paper-round2.pdf},
         volume = {1},
         year = {2021}
    }

@article{cai2025vdd,
  title={Vdd: Varied drone dataset for semantic segmentation},
  author={Cai, Wenxiao and Jin, Ke and Hou, Jinyan and Guo, Cong and Wu, Letian and Yang, Wankou},
  journal={Journal of Visual Communication and Image Representation},
  volume={109},
  pages={104429},
  year={2025},
  publisher={Elsevier}
}

@inproceedings{xu2023side,
  author    = {Xu, Ming and Zhang, Zhen and Wei, Feng and Hu, Han and Bai, Xiang},
  title     = {Side adapter network for open-vocabulary semantic segmentation},
  booktitle = {Proceedings of the IEEE/CVF Conference on Computer Vision and Pattern Recognition},
  pages     = {2945--2954},
  year      = {2023}
}

@inproceedings{xu2022simple,
  author    = {Xu, Ming and Zhang, Zhen and Wei, Feng and Lin, Yixuan and Cao, Yukun and Hu, Han and Bai, Xiang},
  title     = {A simple baseline for open-vocabulary semantic segmentation with pre-trained vision-language model},
  booktitle = {European Conference on Computer Vision},
  pages     = {736--753},
  publisher = {Springer},
  year      = {2022}
}

@article{li2024segearth,
  title={Segearth-ov: Towards training-free open-vocabulary segmentation for remote sensing images},
  author={Li, Kaiyu and Liu, Ruixun and Cao, Xiangyong and Bai, Xueru and Zhou, Feng and Meng, Deyu and Wang, Zhi},
  journal={arXiv preprint arXiv:2410.01768},
  year={2024}
}

@inproceedings{liang2023open,
  author    = {Liang, Feng and Wu, Baitao and Dai, Xinyu and Li, Kuan and Zhao, Yue and Zhang, Han and Zhang, Peng and Vajda, Peter and Marculescu, Daniel},
  title     = {Open-vocabulary semantic segmentation with mask-adapted clip},
  booktitle = {Proceedings of the IEEE/CVF Conference on Computer Vision and Pattern Recognition},
  pages     = {7061--7070},
  year      = {2023}
}

@misc{wu2019detectron2,
  author =       {Yuxin Wu and Alexander Kirillov and Francisco Massa and
                  Wan-Yen Lo and Ross Girshick},
  title =        {Detectron2},
  howpublished = {\url{https://github.com/facebookresearch/detectron2}},
  year =         {2019}
}

@inproceedings{xie2024sed,
  author    = {Xie, Bo and Cao, Jie and Xie, Jing and Khan, Fahad Shahbaz and Pang, Youtao},
  title     = {Sed: A simple encoder-decoder for open-vocabulary semantic segmentation},
  booktitle = {Proceedings of the IEEE/CVF Conference on Computer Vision and Pattern Recognition},
  pages     = {3426--3436},
  year      = {2024}
}

@article{li2024stitchfusion,
  author  = {Li, Bowen and Zhang, Dong and Zhao, Zhengyuan and Gao, Junyu and Li, Xuelong},
  title   = {StitchFusion: Weaving Any Visual Modalities to Enhance Multimodal Semantic Segmentation},
  journal = {arXiv preprint arXiv:2408.01343},
  year    = {2024}
}

@article{chaudhuri2017multilabel,
  title={Multilabel remote sensing image retrieval using a semisupervised graph-theoretic method},
  author={Chaudhuri, Bindita and Demir, Beg{\"u}m and Chaudhuri, Subhasis and Bruzzone, Lorenzo},
  journal={IEEE Transactions on Geoscience and Remote Sensing},
  volume={56},
  number={2},
  pages={1144--1158},
  year={2017},
}

@article{yao2021scale,
  title={Scale-aware detailed matching for few-shot aerial image semantic segmentation},
  author={Yao, Xiwen and Cao, Qinglong and Feng, Xiaoxu and Cheng, Gong and Han, Junwei},
  journal={IEEE Transactions on Geoscience and Remote Sensing},
  volume={60},
  pages={1--11},
  year={2021},
  publisher={IEEE}
}

@article{li2024u3m,
  author  = {Li, Bowen and Zhang, Dong and Zhao, Zhengyuan and Gao, Junyu and Li, Xuelong},
  title   = {U3M: Unbiased Multiscale Modal Fusion Model for Multimodal Semantic Segmentation},
  journal = {arXiv preprint arXiv:2405.15365},
  year    = {2024}
}

@article{chen2017deeplab,
  author    = {Chen, Liang-Chieh and Papandreou, George and Kokkinos, Iasonas and Murphy, Kevin and Yuille, Alan L.},
  title     = {Deeplab: Semantic image segmentation with deep convolutional nets, atrous convolution, and fully connected crfs},
  journal   = {IEEE Transactions on Pattern Analysis and Machine Intelligence},
  volume    = {40},
  number    = {4},
  pages     = {834--848},
  publisher = {IEEE},
  year      = {2017}
}

@article{yu2023convolutions,
  author  = {Yu, Qingyi and He, Jiahao and Deng, Xin and Shen, Xiaohui and Chen, Liang-Chieh},
  title   = {Convolutions die hard: Open-vocabulary segmentation with single frozen convolutional clip},
  journal = {Advances in Neural Information Processing Systems},
  volume  = {36},
  pages   = {32215--32234},
  year    = {2023}
}

@inproceedings{radford2021learning,
  author    = {Radford, Alec and Kim, Jong Wook and Hallacy, Chris and Ramesh, Aditya and Goh, Gabriel and Agarwal, Sandhini and Sastry, Girish and Askell, Amanda and Mishkin, Pamela and Clark, Jack and others},
  title     = {Learning transferable visual models from natural language supervision},
  booktitle = {International Conference on Machine Learning},
  pages     = {8748--8763},
  publisher = {PMLR},
  year      = {2021}
}

@inproceedings{jia2021scaling,
  author    = {Jia, Chen and Yang, Yinfei and Xia, Ye and Chen, Yi-Ting and Parekh, Zarana and Pham, Hieu and Le, Quoc and Sung, Yonghui and Li, Zhifeng and Duerig, Tom},
  title     = {Scaling up visual and vision-language representation learning with noisy text supervision},
  booktitle = {International Conference on Machine Learning},
  pages     = {4904--4916},
  publisher = {PMLR},
  year      = {2021}
}

@inproceedings{long2015fully,
  author    = {Long, Jonathan and Shelhamer, Evan and Darrell, Trevor},
  title     = {Fully convolutional networks for semantic segmentation},
  booktitle = {Proceedings of the IEEE Conference on Computer Vision and Pattern Recognition},
  pages     = {3431--3440},
  year      = {2015}
}

@inproceedings{ronneberger2015u,
  author    = {Ronneberger, Olaf and Fischer, Philipp and Brox, Thomas},
  title     = {U-net: Convolutional networks for biomedical image segmentation},
  booktitle = {Medical Image Computing and Computer-Assisted Intervention--MICCAI 2015: 18th International Conference, Munich, Germany, October 5-9, 2015, Proceedings, Part III},
  pages     = {234--241},
  publisher = {Springer},
  year      = {2015}
}

@article{badrinarayanan2017segnet,
  author    = {Badrinarayanan, Vijay and Kendall, Alex and Cipolla, Roberto},
  title     = {SegNet: A deep convolutional encoder-decoder architecture for image segmentation},
  journal   = {IEEE Transactions on Pattern Analysis and Machine Intelligence},
  volume    = {39},
  number    = {12},
  pages     = {2481--2495},
  publisher = {IEEE},
  year      = {2017}
}

@article{dosovitskiy2020image,
  author  = {Dosovitskiy, Alexey},
  title   = {An image is worth 16x16 words: Transformers for image recognition at scale},
  journal = {arXiv preprint arXiv:2010.11929},
  year    = {2020}
}

@article{xie2021segformer,
  author  = {Xie, Enze and Wang, Wenhai and Yu, Zhiding and Anandkumar, Anima and Alvarez, Jose M. and Luo, Ping},
  title   = {SegFormer: Simple and efficient design for semantic segmentation with transformers},
  journal = {Advances in Neural Information Processing Systems},
  volume  = {34},
  pages   = {12077--12090},
  year    = {2021}
}

@article{hu2024contrastive,
  author  = {Hu, Z. and Gao, J. and Yuan, Y. and Li, X.},
  title   = {Contrastive Tokens and Label Activation for Remote Sensing Weakly Supervised Semantic Segmentation},
  journal = {IEEE Transactions on Geoscience and Remote Sensing},
  year    = {2024}
}

@inproceedings{lin2018deeptongue,
  author    = {Lin, Bo and Xie, Jing and Li, Chang and Qu, Yan},
  title     = {Deeptongue: Tongue segmentation via ResNet},
  booktitle = {2018 IEEE International Conference on Acoustics, Speech and Signal Processing (ICASSP)},
  pages     = {1035--1039},
  publisher = {IEEE},
  year      = {2018}
}

@inproceedings{cheng2022masked,
  author    = {Cheng, Bowen and Misra, Ishan and Schwing, Alexander G. and Kirillov, Alexander and Girdhar, Rohit},
  title     = {Masked - attention mask transformer for universal image segmentation},
  booktitle = {Proceedings of the IEEE/CVF Conference on Computer Vision and Pattern Recognition},
  pages     = {1290--1299},
  publisher = {IEEE},
  year      = {2022}
}

@article{cheng2021per,
  author  = {Cheng, Bowen and Schwing, Alexander and Kirillov, Alexander},
  title   = {Per - pixel classification is not all you need for semantic segmentation},
  journal = {Advances in Neural Information Processing Systems},
  volume  = {34},
  pages   = {17864--17875},
  year    = {2021}
}

@inproceedings{zhou2022extract,
  author    = {Zhou, Chang and Loy, Chen Change and Dai, Bo},
  title     = {Extract free dense labels from CLIP},
  booktitle = {European Conference on Computer Vision},
  pages     = {696--712},
  publisher = {Springer},
  year      = {2022}
}

@inproceedings{jiao2025collaborative,
  author    = {Jiao, Shilei and Zhu, Hongtu and Huang, Junjie and Zhao, Yue and Wei, Yunchao and Shi, Hujun},
  title     = {Collaborative vision - text representation optimizing for open - vocabulary segmentation},
  booktitle = {Proceedings of the European Conference on Computer Vision},
  pages     = {399--416},
  year      = {2025}
}

@article{xie2024multi,
  author  = {Xie, Bo and Cao, Jie and Anwer, Raja M. and Xie, Jing and Nie, Jin and Yang, Ailong and Pang, Youtao},
  title   = {Multi - query and multi - level enhanced network for semantic segmentation},
  journal = {Pattern Recognition},
  volume  = {156},
  pages   = {110777},
  publisher = {Elsevier},
  year    = {2024}
}

@inproceedings{zhu2024saswot,
  author    = {Zhu, Chen and Li, Liang and Wu, Yanjie and Sun, Zhen},
  title     = {Saswot: Real - time semantic segmentation architecture search without training},
  booktitle = {Proceedings of the AAAI Conference on Artificial Intelligence},
  volume    = {38},
  number    = {7},
  pages     = {7722--7730},
  year      = {2024}
}

@article{xu2024generalization,
  author  = {Xu, Wenhan and Wang, Chen and Feng, Xin and Xu, Runze and Huang, Lei and Zhang, Zhen and Guo, Lei and Xu, Shuaicheng},
  title   = {Generalization boosted adapter for open - vocabulary segmentation},
  journal = {IEEE Transactions on Circuits and Systems for Video Technology},
  publisher = {IEEE},
  year    = {2024}
}

@inproceedings{shan2024open,
  author    = {Shan, Xudong and Wu, Di and Zhu, Guorong and Shao, Yong and Sang, Nong and Gao, Changxin},
  title     = {Open - vocabulary semantic segmentation with image embedding balancing},
  booktitle = {Proceedings of the IEEE/CVF Conference on Computer Vision and Pattern Recognition},
  pages     = {28412--28421},
  year      = {2024}
}

\end{document}